\documentclass{article}

\usepackage{arxiv}

\usepackage[utf8]{inputenc} 
\usepackage[T1]{fontenc}    
\usepackage{hyperref}       
\usepackage{url}            
\usepackage{booktabs}       
\usepackage{amsfonts}       
\usepackage{nicefrac}       
\usepackage{microtype}      
\usepackage{lipsum}
\usepackage{graphicx}
\usepackage{xcolor}
\usepackage[square,numbers]{natbib}  
\usepackage{textcomp, gensymb} 
\usepackage{subcaption}
\usepackage{mathtools}
\graphicspath{ {./Figs/} }

\title{A Two‑Stage Transformer Framework for Temporal Localization of Distracted Driver Behaviors}

\author{
 Gia-Bao Doan \\
  Department of Information Technology\\
  FPT University\\
  Ho Chi Minh City 700000, Vietnam\\
  \texttt{tonybdg2061@gmail.com}\\
 \And
 Nam-Khoa Huynh \\
  Department of Information Technology\\
  FPT University\\
  Ho Chi Minh City 700000, Vietnam\\
  \texttt{khoanamhuynh2004@gmail.com}\\
 \And 
 Minh-Nhat-Huy Ho \\
  Department of Information Technology\\
  FPT University\\
  Ho Chi Minh City 700000, Vietnam\\
  \texttt{nhathuyho04.work@gmail.com}\\
 \And 
 Khanh-Thanh-Khoa Nguyen \\
  Department of Information Technology\\
  FPT University\\
  Ho Chi Minh City 700000, Vietnam\\
  \texttt{ngkthkhoa@gmail.com}\\
 \And
 Thi-Thu-Hien Pham \\
  Faculty of Electronics Technology \\
  Industrial University of Ho Chi Minh City \\ 
  Ho Chi Minh City 700000, Vietnam \\
  \texttt{phamthithuhien@iuh.edu.vn}\\
 \And 
 Thanh-Hai Le \\
  School of Computer Science and Engineering\\
  The Saigon International University\\
  Ho Chi Minh City 700000, Vietnam\\
  \texttt{lethanhhai@siu.edu.vn}\\
}

\begin{document}
\maketitle
\begin{abstract}
	Identifying hazardous driver behaviors from in-cabin video streams is vital for enhancing road safety and automating traffic violation detection. However, existing temporal action localization (TAL) techniques frequently struggle to balance localization accuracy with computational efficiency. In this work, we present a robust two-stage TAL framework tailored for driver monitoring scenarios, designed for post-trip fleet audits and periodic transportation safety inspections. The proposed pipeline pairs a VideoMAE-based feature extractor with an Augmented Self-Mask Attention (AMA) localization head, enhanced by a 1D Spatial Pyramid Pooling-Fast (SPPF) module to effectively capture multi-scale temporal context across heterogeneous action durations. 
	
	Our evaluation reveals a clear accuracy--efficiency trade-off across backbone scales. At the feature extraction stage, a massive ViT-Giant backbone yields powerful spatiotemporal representations, achieving $88.09\%$ Top-1 test accuracy. Conversely, a distilled ViT-Base variant serves as a highly efficient alternative, delivering a competitive $82.55\%$ accuracy while drastically reducing computational overhead ($101.85$ GFLOPs/segment vs. $1584.06$ GFLOPs/segment for Giant). For the downstream localization task, the integration of the SPPF neck consistently drives performance gains across all settings. Notably, the ViT-Giant + SPPF configuration achieves a peak mean Average Precision (mAP) of $92.67\%$, while the lightweight ViT-Base counterpart maintains strong, deployment-ready results.
\end{abstract}


\section{Introduction}
\label{sec:intro}
Ensuring driver attentiveness is critical for road safety, as human error remains a leading cause of global traffic accidents. Driven by in-cabin cameras, automated driver monitoring has become feasible, yet reliably recognizing subtle, brief, or partially hidden distraction behaviors (e.g., mobile-phone use, eating) remains highly challenging due to variable illumination, frequent occlusions, and severe inter-driver variability \cite{bouhsissin2023driver, hurts2011distracted, eraqi2019driver, jha2021multimodal}. Classical CNN-based frameworks and ensemble architectures \cite{kumar2021computer} struggle with these continuous, fine-grained tasks because they operate primarily on spatial cues and lack adequate temporal modeling capacity.

Similarly, traditional temporal action localization (TAL) methods that rely on sliding windows or anchors \cite{shou2016temporal, gao2017cascaded, lin2017single} fail to accommodate highly variable action durations within long, untrimmed driving videos. This frequently causes short actions to be missed and extended behaviors to be fragmented. Recent surveys \cite{fu2024advancements} confirm that standard CNN- and LSTM-based approaches lack the representational power needed to model long-range dependencies and multi-scale contexts in naturalistic driving data.

While transformer-based architectures—ranging from early spatio-temporal attention designs \cite{bertasius2021space, han2021transformer} to specialized TAL frameworks like ActionFormer \cite{zhang2022actionformer} and TriDet \cite{shi2023tridet}—have improved boundary precision, they still rely predominantly on local-window attention. This limits their ability to capture global temporal context across distant video segments when actions occur sparsely. Concurrently, self-supervised pretraining via masked autoencoders, such as VideoMAE \cite{tong2022videomae} and VideoMAE V2 \cite{wang2023videomae}, has demonstrated strong resilience to noisy conditions by learning robust spatio-temporal features.

Despite these architectural and self-supervised advancements, existing frameworks still fall short of the temporal boundary precision required for real-world deployment. To bridge this gap and better capture long-range dependencies under realistic driving conditions, we adopt and extend the Augmented Self-Mask Attention framework \cite{zhang2024augmented} to precisely localize fine-grained distracted behaviors.

\section{Related Works}
\label{sec:relatedworks}
\subsection{Video-based Action Recognition and Representation Learning}
Video recognition serves as the foundation for temporal action localization (TAL). Early convolutional architectures, including 3D CNNs and segment-based frameworks \cite{bouhsissin2023driver, fu2024advancements}, primarily targeted clip-level classification. Recently, transformer-based models like TimeSformer \cite{bertasius2021space} and Transformer-in-Transformer \cite{han2021transformer} extended self-attention to capture long-range spatio-temporal dependencies. Concurrently, self-supervised masked video modeling—such as MaskedFeat \cite{wei2022masked}, MAE-ST \cite{feichtenhofer2022masked}, VideoMAE \cite{tong2022videomae}, and VideoMAE V2 \cite{wang2023videomae}—demonstrated that reconstructing heavily masked tokens yields robust representations. 

However, these frameworks are designed for recognition rather than boundary localization. Furthermore, scaling up to large ViT backbones introduces severe computational costs, making them impractical for resource-constrained in-cabin driver monitoring. To balance accuracy and efficiency, we build on the VideoMAE family \cite{tong2022videomae, wang2023videomae} by utilizing a ViT-Base backbone distilled from a larger ViT-Giant model, preserving powerful representation quality at a lower computational footprint.

\subsection{Temporal Action Localization}
Temporal Action Localization (TAL) identifies action categories and their precise temporal boundaries in untrimmed videos. Early sliding-window methods (e.g., S-CNN \cite{shou2016temporal}, TURN-TAP \cite{gao2017turn}, CBR \cite{gao2017cascaded}) and two-stage proposal-based pipelines \cite{bouhsissin2023driver, fu2024advancements} suffer from high computational redundancy and struggle with variable-duration actions. Single-stage anchor-based detectors like SSAD \cite{lin2017single} and GTAN \cite{long2019gaussian} bypass proposals but rely on heuristic anchor configurations that generalize poorly.

State-of-the-art transformer-based TAL models, such as ActionFormer \cite{zhang2022actionformer} and TriDet \cite{shi2023tridet}, leverage self-attention and refined boundary modeling. However, their reliance on long input sequences and substantial GPU memory limits their deployment in constrained environments or high-throughput post-trip auditing. To address these efficiency and multi-scale modeling bottlenecks, the Augmented Self-Mask Attention (AMA) transformer \cite{zhang2024augmented} was introduced for naturalistic driving. Our work integrates AMA with efficient VideoMAE-based features in a unified, two-stage pipeline for temporal driver behavior localization.

\subsection{Driver Behavior Recognition}
Driver behavior recognition is essential for advanced driver assistance systems (ADAS) \cite{bouhsissin2023driver, fu2024advancements}. Grounded in human factors research \cite{hurts2011distracted}, early deep learning approaches leveraged CNN ensembles \cite{eraqi2019driver} or genetic algorithms \cite{kumar2021computer} for image- or short-clip state classification. While effective for coarse distraction detection, these methods fail to capture extended, subtle, or continuous temporal behaviors.

While multimodal datasets like MDMD \cite{jha2021multimodal} incorporate IMU, CAN bus, and biometric signals, such systems are costly and complex to deploy at scale. Conversely, the AI City Challenge Track 3 dataset \cite{wang20248th} and our self-collected dataset operate under a video-only paradigm using synchronized in-cabin cameras. Aligning with this practical constraint, our framework pairs a distilled VideoMAE backbone with an AMA-driven transformer to deliver high-throughput, computationally efficient temporal localization for offline naturalistic driving video analysis.

\section{Materials and Methods}
\label{sec:materials}
\subsection{Dataset}
We evaluate our approach using the \textbf{AI City Challenge 2024 Track 3 dataset} \cite{wang20248th}, designed for multi-angle, in-cabin driver distraction detection. It contains 594 video clips ($\approx$90 hours) from 99 drivers, each performing 16 distinct tasks (e.g., phone use, eating) from three synchronized camera views. The sequence is performed under normal conditions and with appearance blocks (e.g., hats, sunglasses) to maximize variability as described in Table \ref{tab:actions}. 

\begin{figure}[h]
	\centering	
	\begin{subfigure}[b]{0.32\textwidth}
		\includegraphics[width=\textwidth]{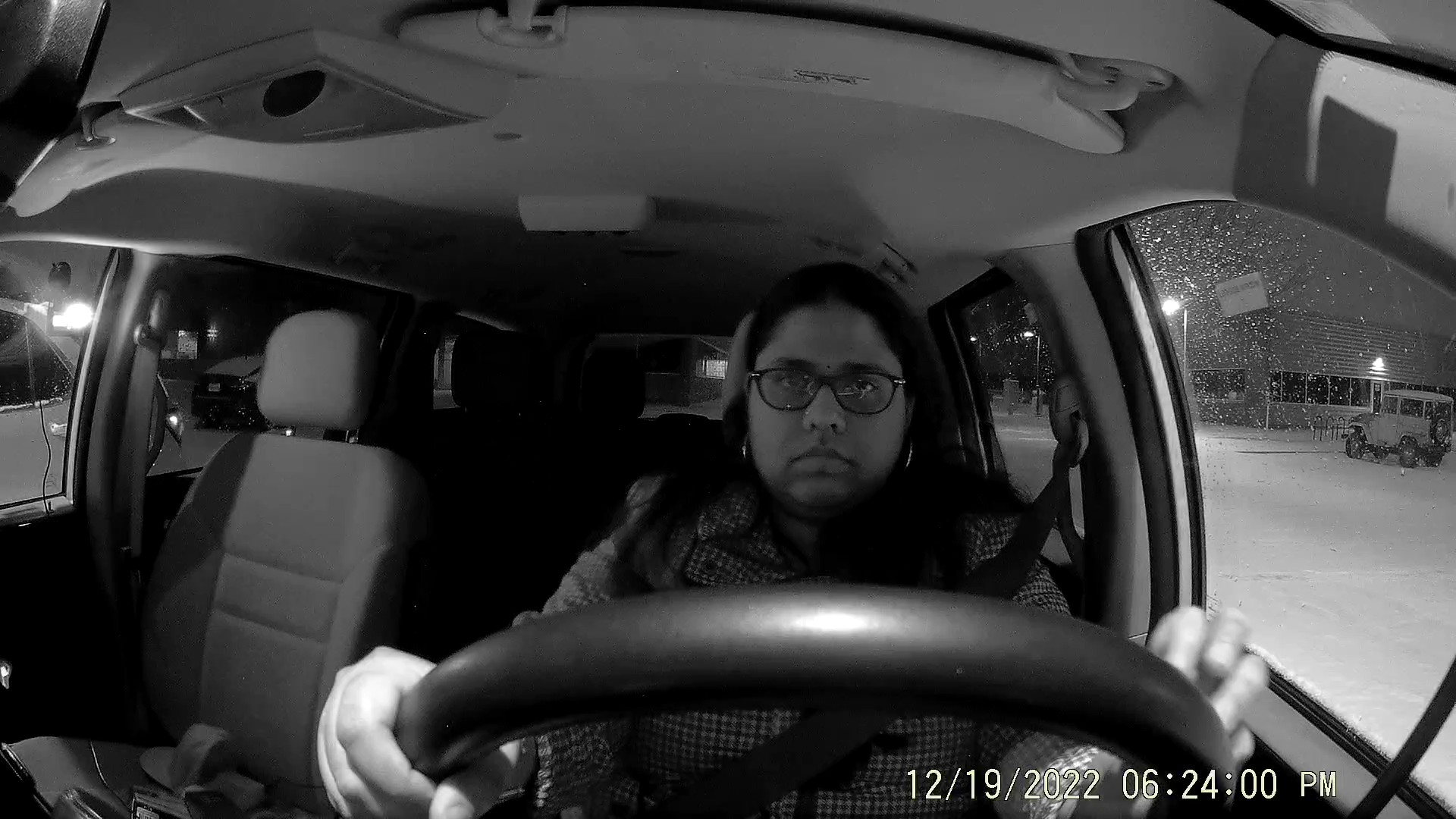}
	\end{subfigure}
	\begin{subfigure}[b]{0.32\textwidth}
		\includegraphics[width=\textwidth]{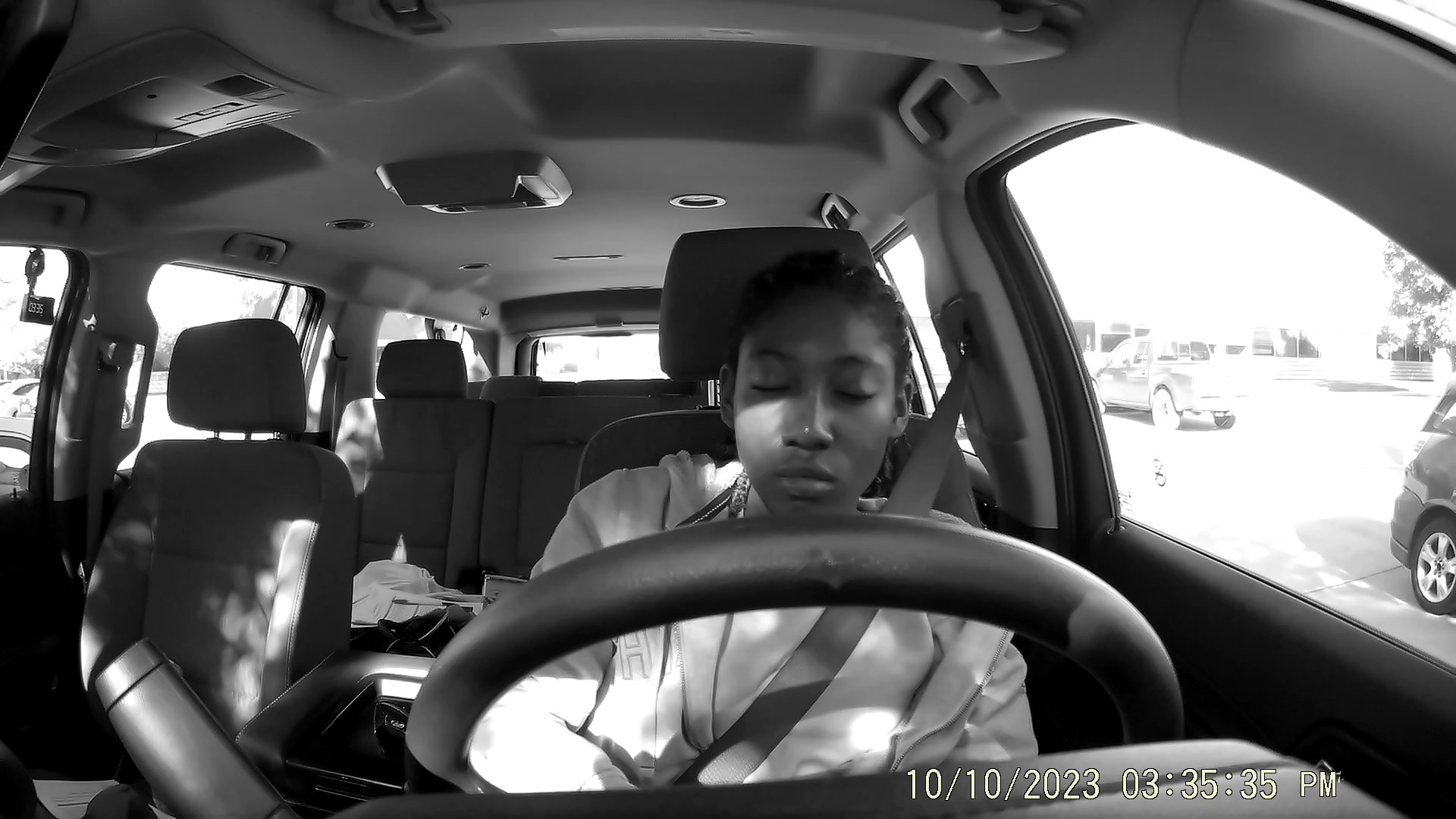}
	\end{subfigure}
	\begin{subfigure}[b]{0.32\textwidth}
		\includegraphics[width=\textwidth]{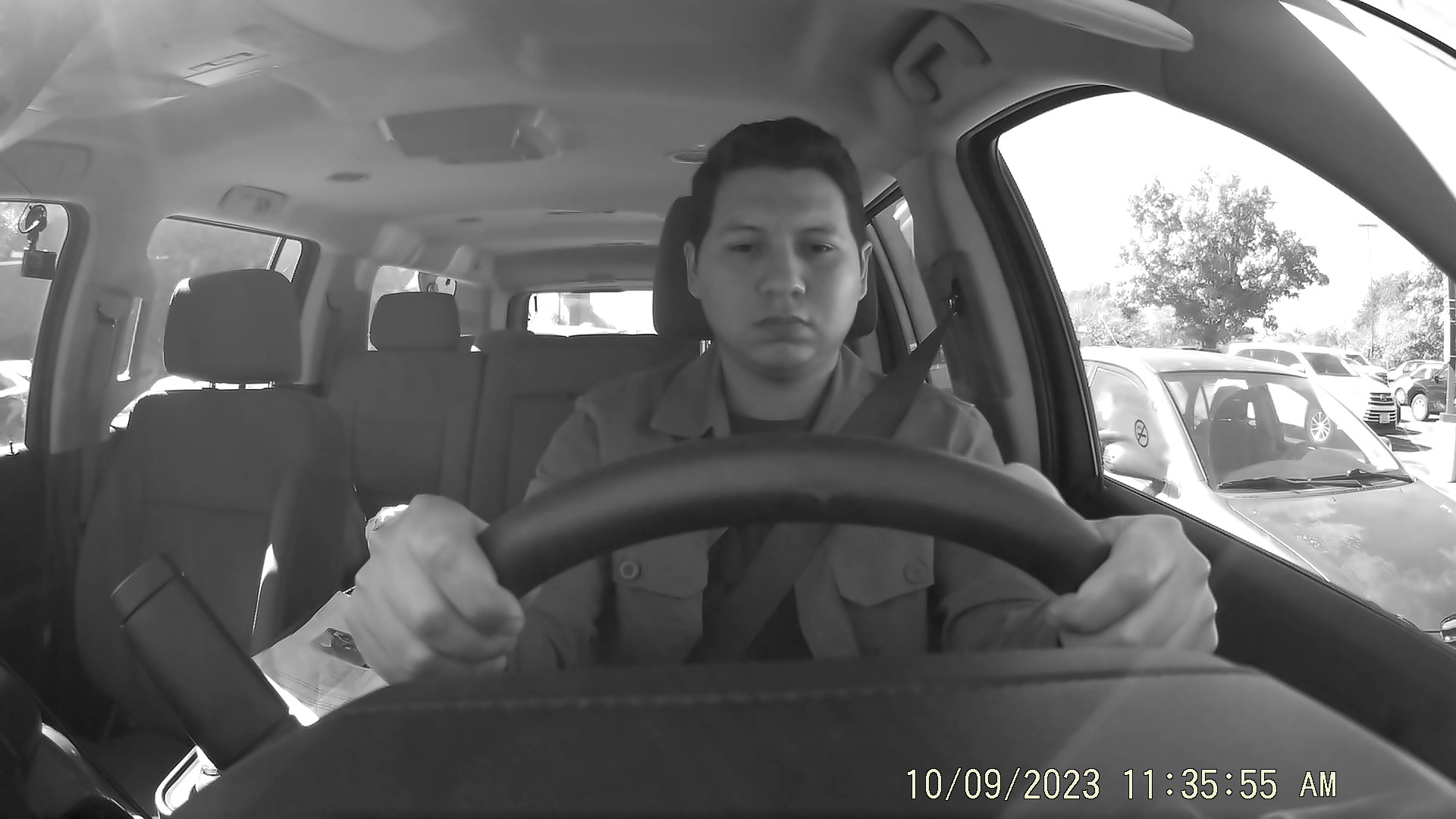}
	\end{subfigure}
	
	\begin{subfigure}[b]{0.32\textwidth}
		\includegraphics[width=\textwidth]{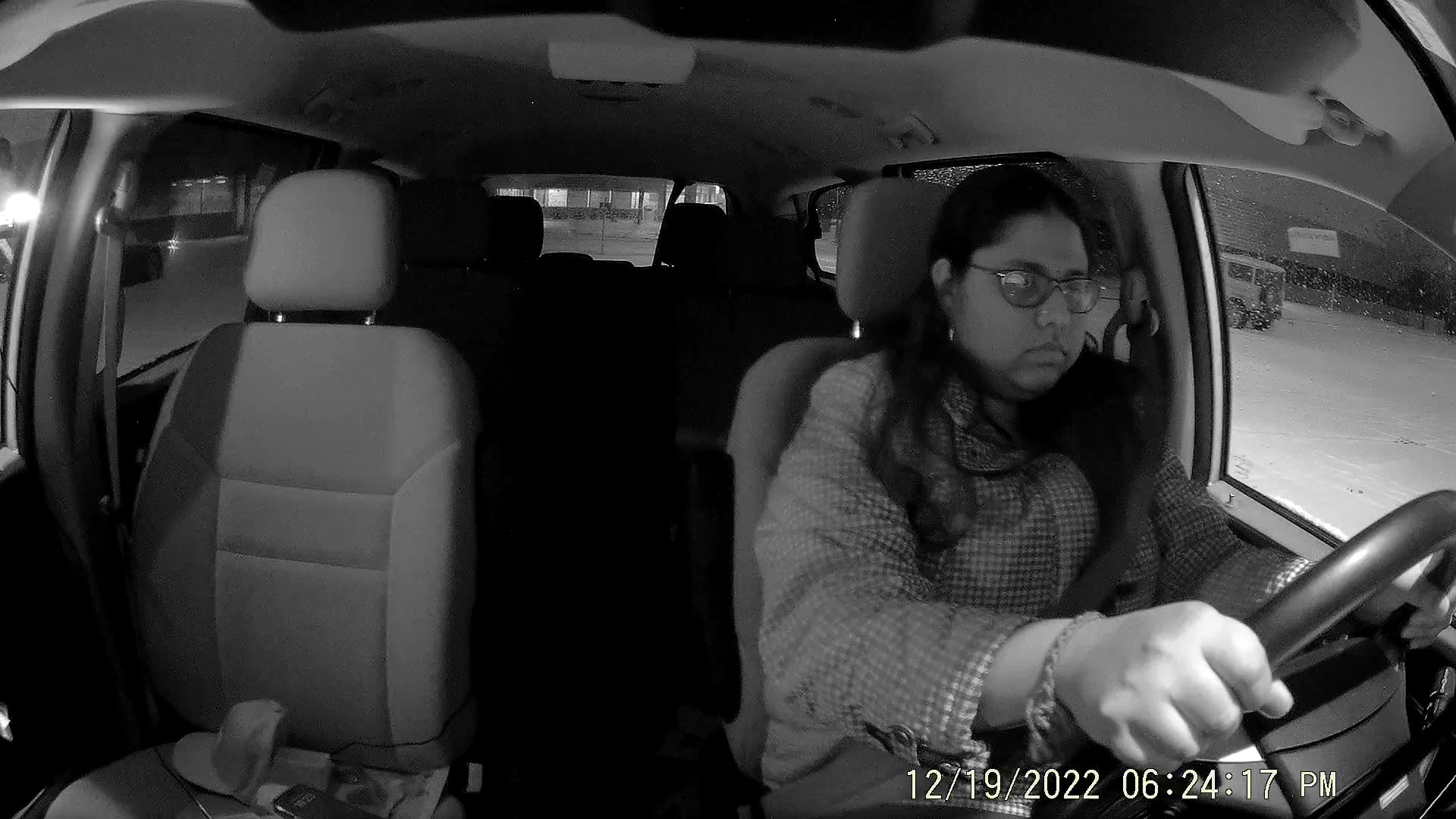}
	\end{subfigure}
	\begin{subfigure}[b]{0.32\textwidth}
		\includegraphics[width=\textwidth]{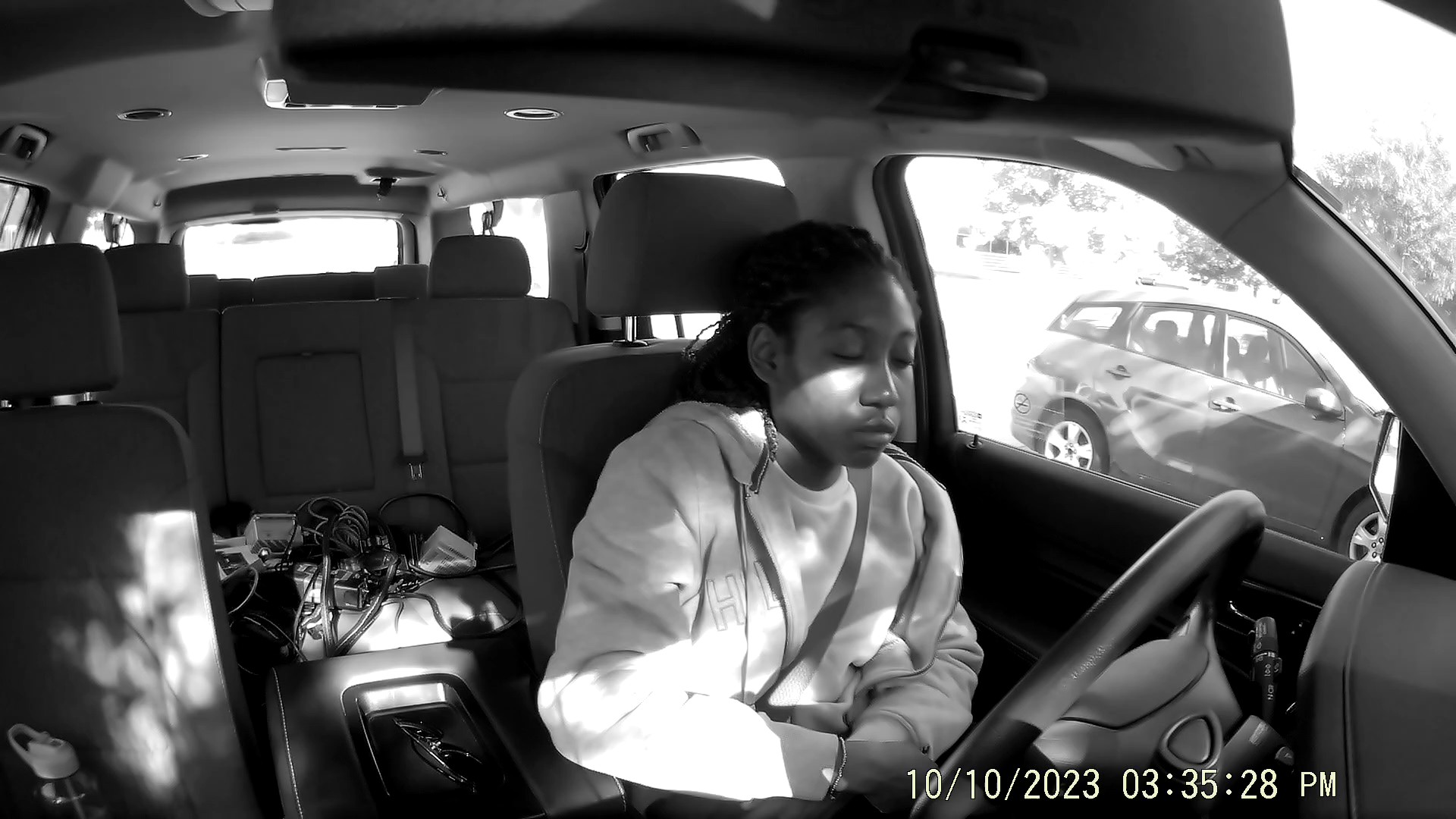}
	\end{subfigure}
	\begin{subfigure}[b]{0.32\textwidth}
		\includegraphics[width=\textwidth]{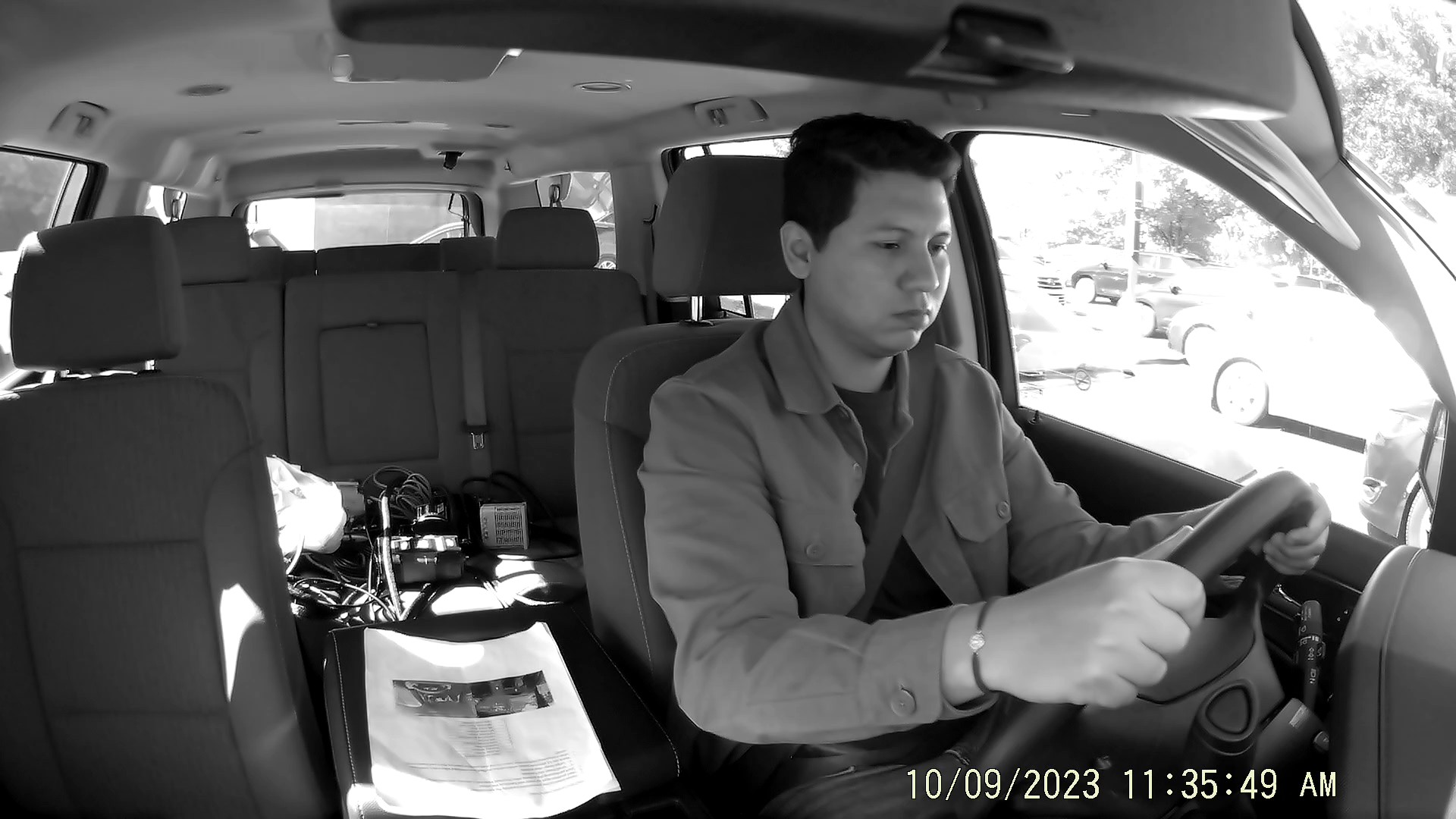}
	\end{subfigure}
	
	\begin{subfigure}[b]{0.32\textwidth}
		\includegraphics[width=\textwidth]{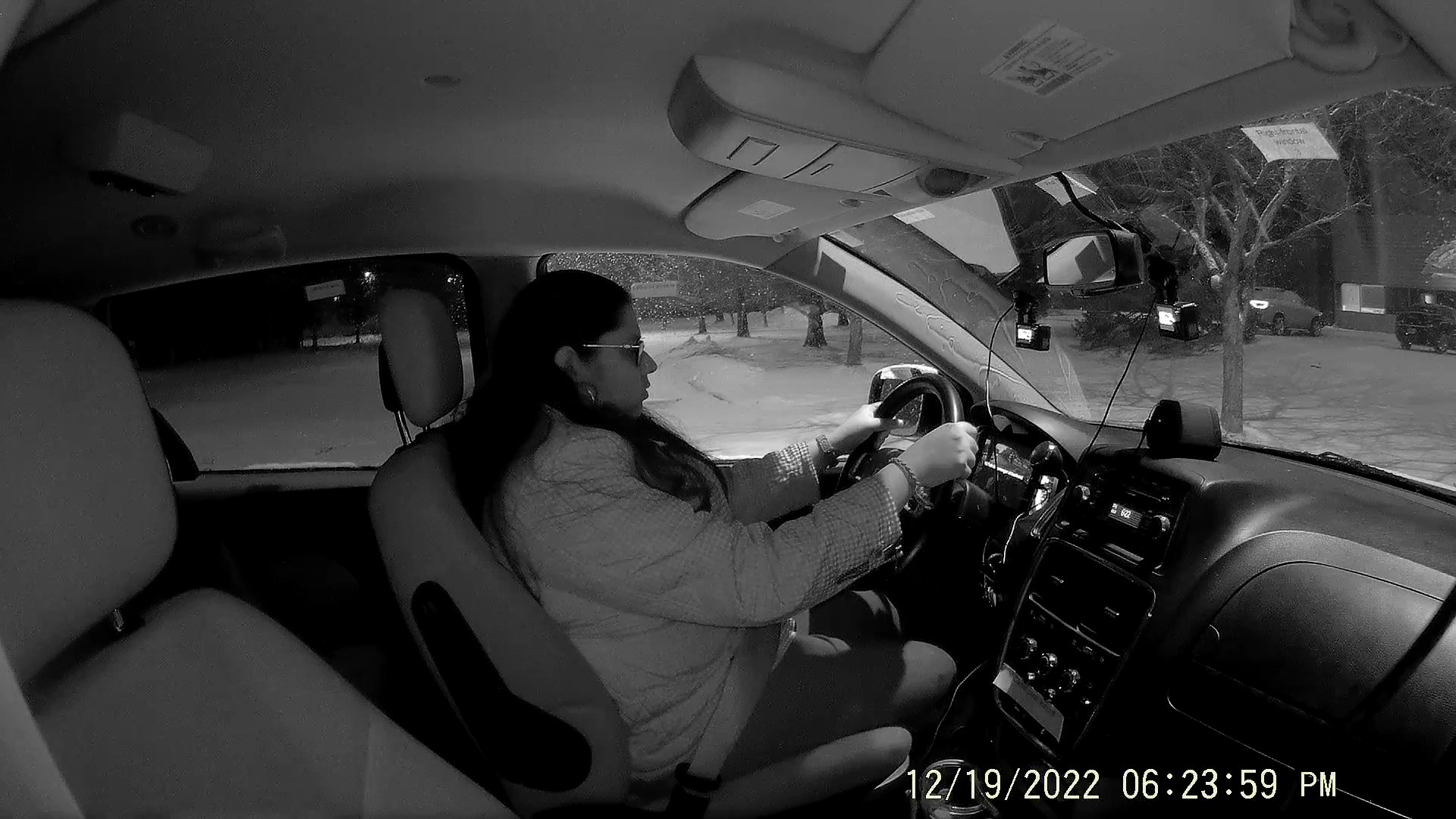}
	\end{subfigure}
	\begin{subfigure}[b]{0.32\textwidth}
		\includegraphics[width=\textwidth]{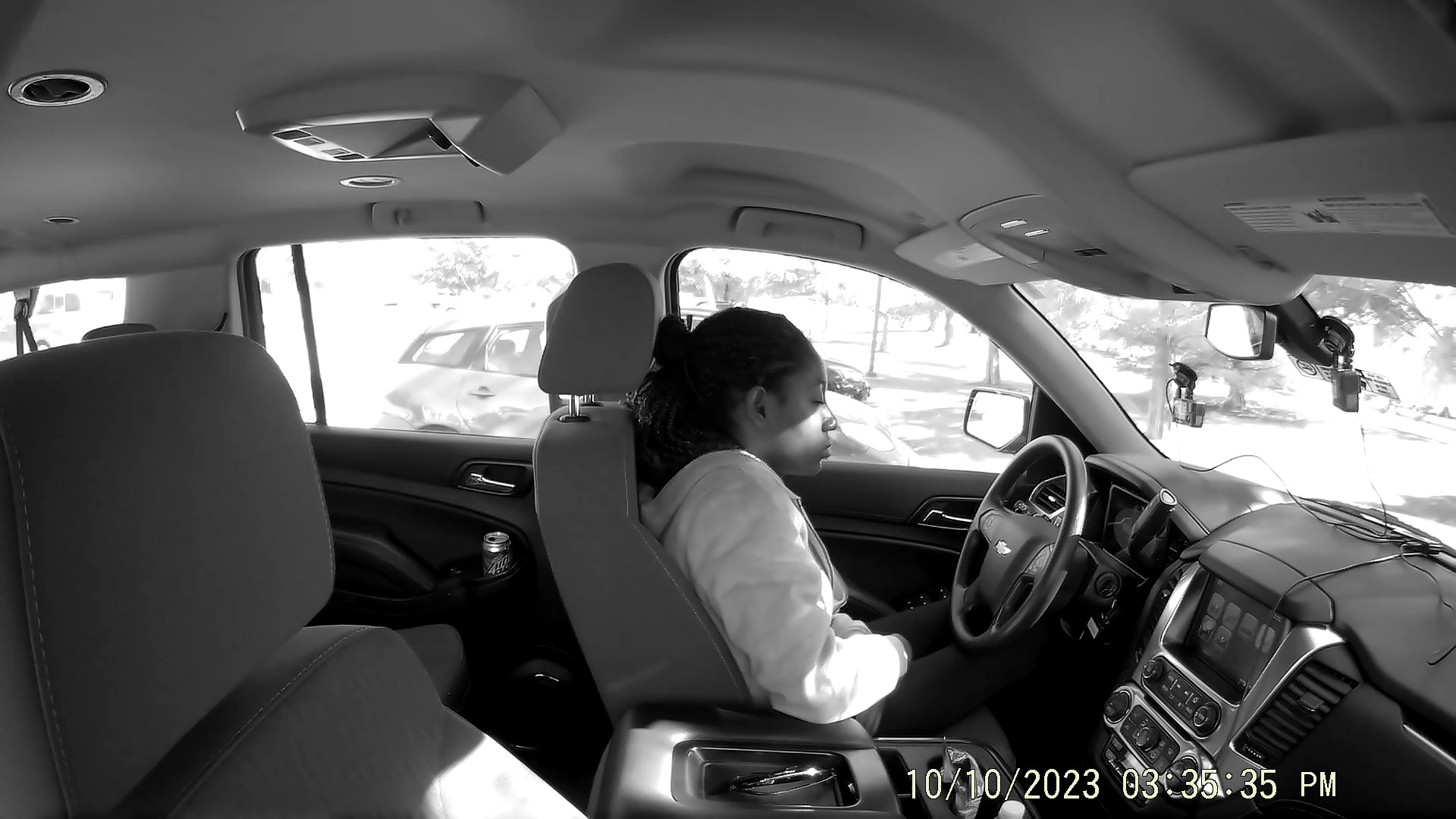}
	\end{subfigure}
	\begin{subfigure}[b]{0.32\textwidth}
		\includegraphics[width=\textwidth]{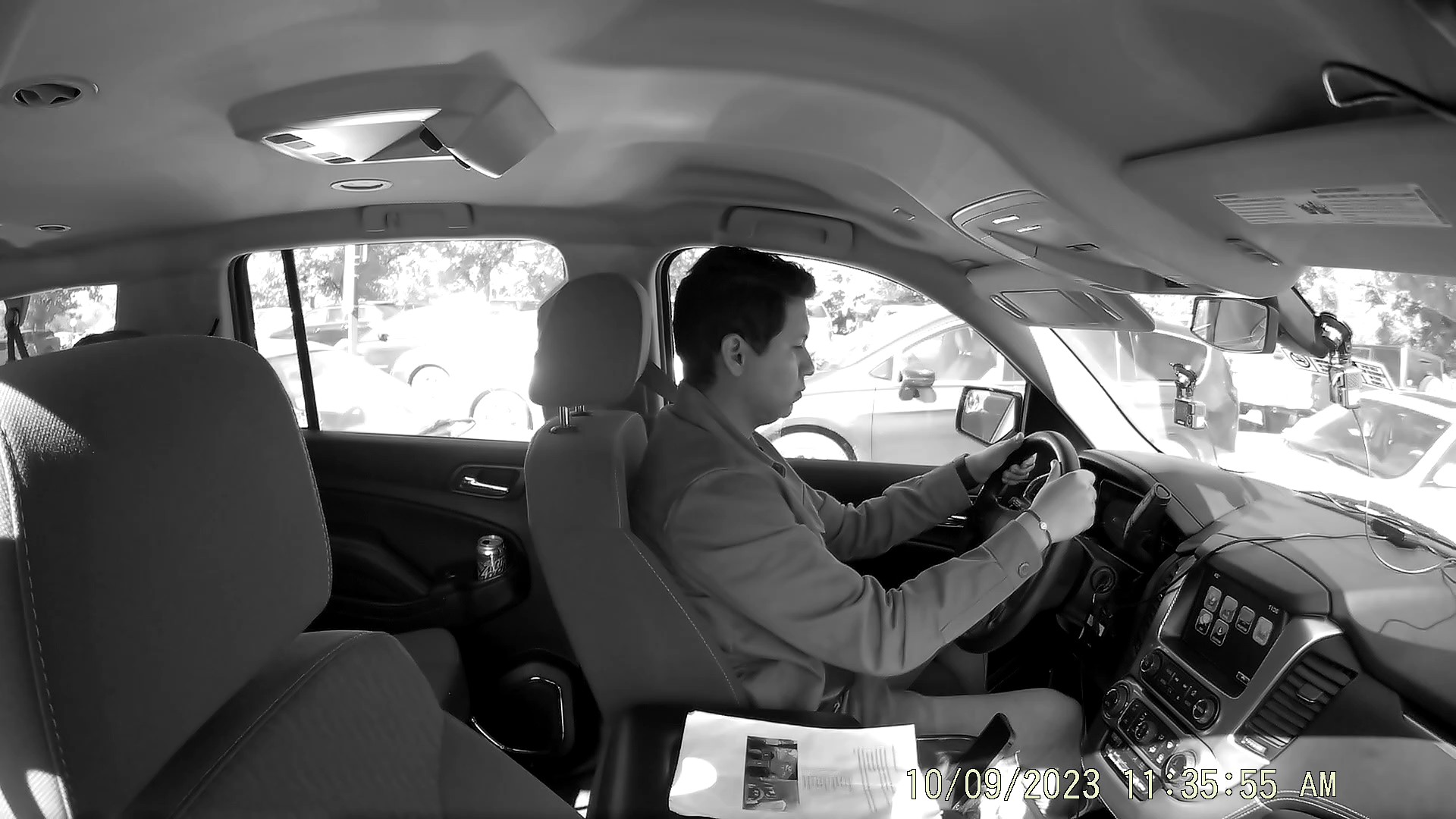}
	\end{subfigure}
	\caption[AI City Challenge 2024 Track 3 dataset overview]{Overview of the AI City Challenge 2024 Track 3 dataset, showing multiple drivers, camera angles, and labeled distraction tasks.}
	\label{fig:original_9_samples}
\end{figure}

The dataset is divided into three subsets based on driver splits: A1 (69 drivers with ground-truth temporal labels for training/validation), A2 (15 drivers without labels for public leaderboard evaluation), and B (15 drivers reserved for final testing) \cite{wang20248th}. Figure \ref{fig:original_9_samples} shows some samples in the AI City Challenge 2024 Track 3 dataset. 

The official evaluation metric is the average activity overlap score. For a ground-truth activity $g$ (start $g_s$, end $g_e$) and a predicted activity $p$ (start $p_s$, end $p_e$) of the same class, a match is valid if $p_s \in [g_s-10\text{s}, g_s+10\text{s}]$ and $p_e \in [g_e-10\text{s}, g_e+10\text{s}]$. The overlap score is the temporal Intersection-over-Union (IoU) of $g$ and $p$, with unmatched instances scoring 0 \cite{wang20248th}.

\subsection{Implementation and Preprocessing Details}
The framework is implemented in PyTorch and evaluated on a cloud server equipped with 64~GB RAM and a single NVIDIA RTX A5000 GPU (24~GB VRAM). 

Following \cite{zhang2024augmented, dong2023multiattention}, raw videos undergo a two-stage preprocessing pipeline:
\begin{itemize}
	\item \textbf{Spatial Stabilization:} YOLOv5 is applied to detect the human body across all frames. To mitigate background shaking from variable bounding boxes, the frame with the largest human detection area defines a fixed cropping window applied globally across the video. This stabilizes the driver position and filters environmental noise.
	\item \textbf{Temporal Segmentation:} Automated scripts parse annotation files to slice untrimmed videos into individual action clips based on ground-truth timestamps. These segments are split into training/validation sets and compiled into JSON metadata files containing boundaries, labels, and properties to streamline feature extraction.
\end{itemize}

\begin{table}[h!]
	\centering
	\caption{List of distracted actions in the AI City Challenge Track 3 dataset}
	\begin{tabular}{|c|l|}
		\hline
		\textbf{Action ID} & \textbf{Distracted Action} \\ \hline
		1  & Drinking \\ \hline
		2  & Phone Call (right hand) \\ \hline
		3  & Phone Call (left hand) \\ \hline
		4  & Eating \\ \hline
		5  & Text (right hand) \\ \hline
		6  & Text (left hand) \\ \hline
		7  & Reaching behind \\ \hline
		8  & Adjust control panel \\ \hline
		9  & Pick up from floor (Driver) \\ \hline
		10 & Pick up from floor (Passenger) \\ \hline
		11 & Talk to passenger (right) \\ \hline
		12 & Talk to passenger (backseat) \\ \hline
		13 & Yawning \\ \hline
		14 & Hand on head \\ \hline
		15 & Singing or dancing with music \\ \hline
		16 & Normal driving \\ \hline
	\end{tabular}
	\label{tab:actions}
\end{table}

\subsection{Methodology}
\subsubsection{Overall Architecture}
\begin{figure}[h]
	\centering
	\resizebox{\textwidth}{!}{%
		\includegraphics[width=1.2\textwidth]{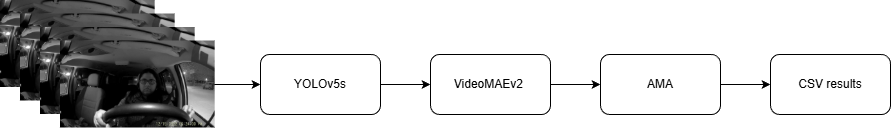}
	}
	\caption{Pipeline of the proposed method}
	\label{fig:pipeline}
\end{figure}

Our temporal action localization (TAL) system uses a two-stage pipeline consisting of three components: (1) driver ROI extraction via YOLOv5s, (2) VideoMAE-based chunk feature extraction, and (3) AMA-driven temporal localization. The overall architecture is shown in Figure \ref{fig:pipeline}.

\subsubsection{ROI Extraction}
To isolate driver behavior and eliminate background motion, we employ YOLOv5s \cite{yolov5}, which provides an optimal speed--accuracy trade-off for real-time and resource-constrained environments \cite{terven2023comprehensive, horvat2022comparative, bao2024weather}. Pretrained on COCO (class 0: person), the detector filters bounding boxes with a confidence threshold $\ge 0.5$. In multi-person cases, the box with the largest spatial area is selected to isolate the primary driver. 

\begin{figure}[h]
	\centering
	\resizebox{\textwidth}{!}{
		\includegraphics[width=1.2\textwidth]{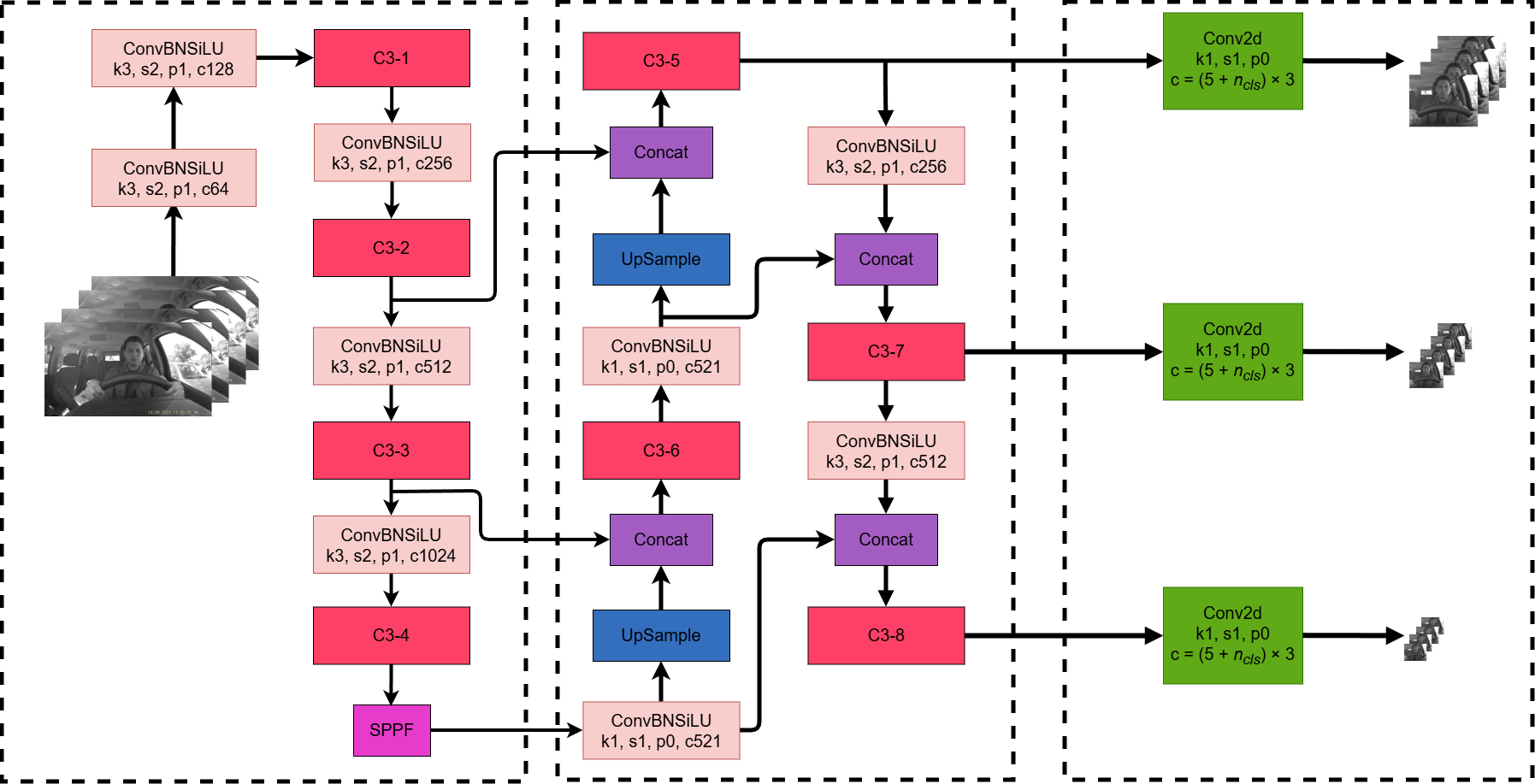}
	}     
	\label{fig:yolov5_architecture}
	\caption{YOLOv5's Architecture}
\end{figure}

The frame-level driver region is cropped and normalized using standard ImageNet statistics. This offline preprocessing step stabilizes the driver position, reduces downstream computational complexity, and generates person-centered video sequences without breaking temporal consistency.

\begin{figure}[h]
	\centering
	\includegraphics[scale=0.5]{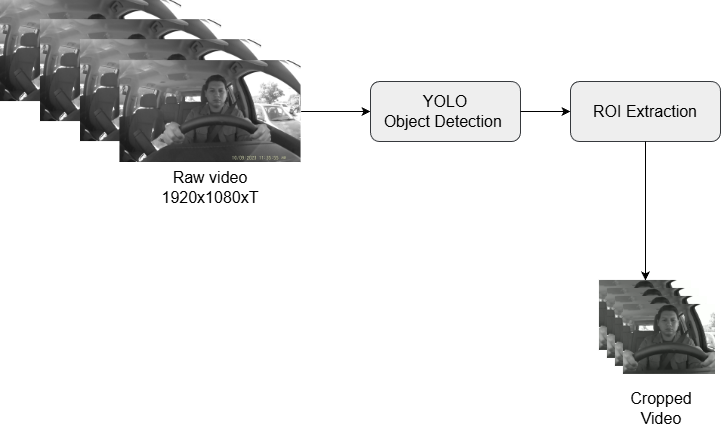}
	\caption{Preprocessing details}
	\label{fig:detailed_pipeline}
\end{figure}

\subsubsection{Feature Extraction}
\label{subsec:Feature_extraction}
We leverage self-supervised video representation learning via masked autoencoders \cite{dosovitskiy2020image, feichtenhofer2022masked, wei2022masked}. VideoMAE \cite{tong2022videomae} uses an asymmetric encoder--decoder with a $\approx 90\%$ masking ratio, while VideoMAE V2 \cite{wang2023videomae} scales this via dual-masking. 

\begin{figure}[h]
	\centering
	\includegraphics[width=0.6\textwidth]{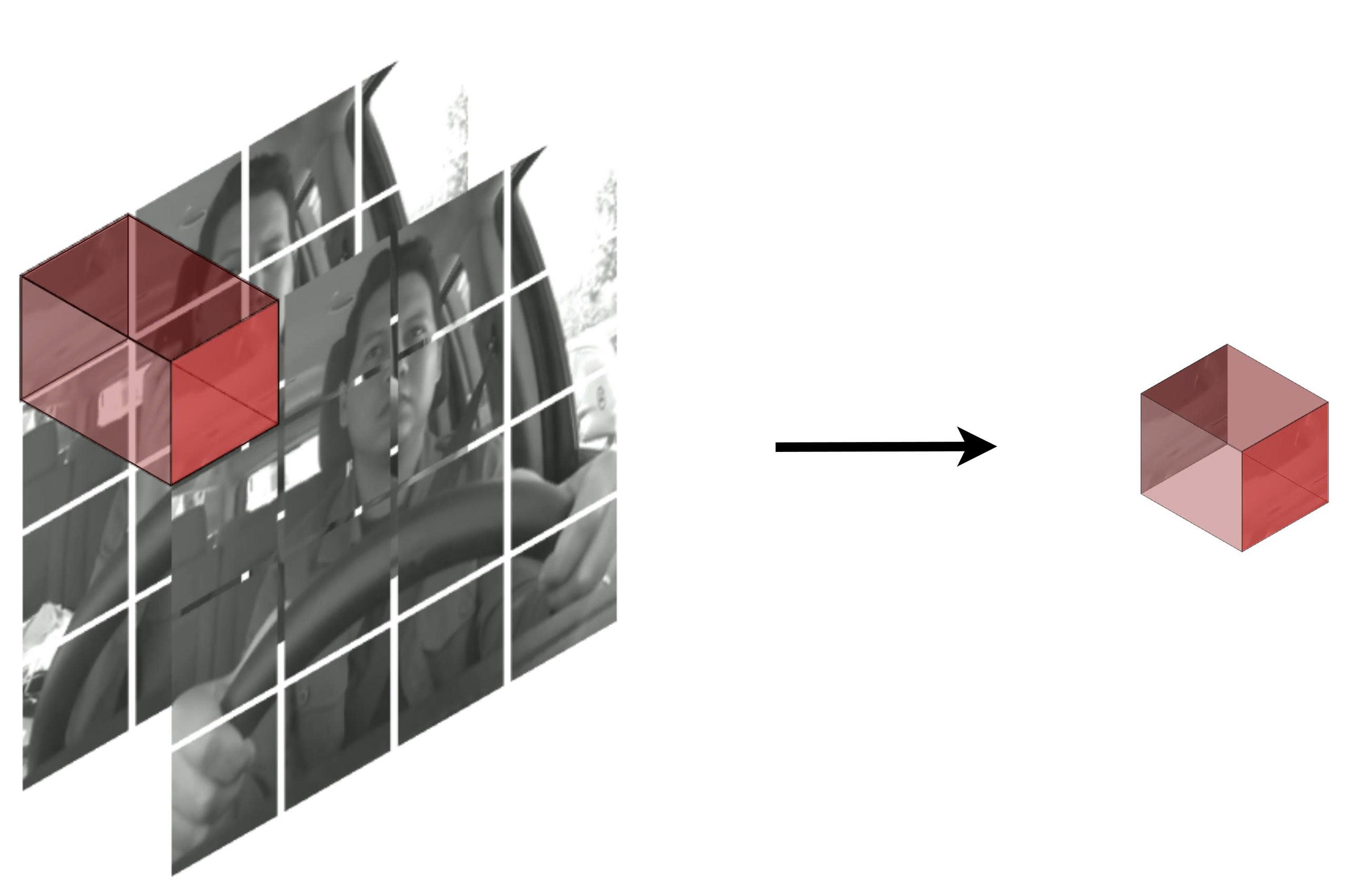}
	\caption{Tubelet aggregation structure.}
	\label{fig:Tubelet}
\end{figure}

An input clip $\mathbf{I} \in \mathbb{R}^{C \times T \times H \times W}$ is tokenized into spatiotemporal tubelets (Figure~\ref{fig:Tubelet}) via 3D convolutions:
\[
\mathbf{T} = \Phi_{emb}(\mathbf{I}) = \{ \mathbf{T}_i \}_{i=1}^{N}.
\]
The visible, unmasked subset $\mathbf{T}^u = \{ \mathbf{T}_i \mid i \notin \mathcal{M}(\rho) \}$ is processed by a Vision Transformer encoder using joint space--time attention \cite{bertasius2021space, han2021transformer}:
\[
\mathbf{Z} = \Phi_{enc}(\mathbf{T}^u).
\]
Pretraining minimizes the mean-squared error over masked tubelets:
\[
\mathcal{L}_{MAE} = \frac{1}{\rho N} \sum_{i \in \mathcal{M}(\rho)} \left\lVert \mathbf{I}_i - \hat{\mathbf{I}}_i \right\rVert_2^2.
\]

For deployment, we adopt a transfer learning strategy rather than reconstruction pretraining, utilizing official VideoMAE V2 checkpoints pretrained on UnlabeledHybrid and fine-tuned on Kinetics-710 (K710). We evaluate two fine-tuned configurations:
\begin{itemize}
	\item \textbf{ViT-Giant}: Checkpoint \texttt{vit\_g\_hybrid\_pt\_1200e\_k710\_ft.pth} ($D=1408$).
	\item \textbf{ViT-Base (distilled)}: Checkpoint \texttt{vit\_b\_k710\_dl\_from\_giant.pth} ($D=768$).
\end{itemize}

During domain-specific fine-tuning on the 16-class AI City Challenge dataset, the reconstruction decoder is removed. Tubelet features are mean-pooled spatiotemporally into a vector $\mathbf{f} \in \mathbb{R}^D$ and mapped to logits via a linear head: $\mathbf{y} = \text{Head}(\mathbf{f}) \in \mathbb{R}^{16}$. Optimization switches to cross-entropy loss:
\[
\mathcal{L}_{\text{cls}} = -\frac{1}{N} \sum_{i=1}^{N} \log P(y_i \mid \mathbf{x}_i).
\]

\begin{figure}[h]
	\centering
	\includegraphics[width=1.1\textwidth]{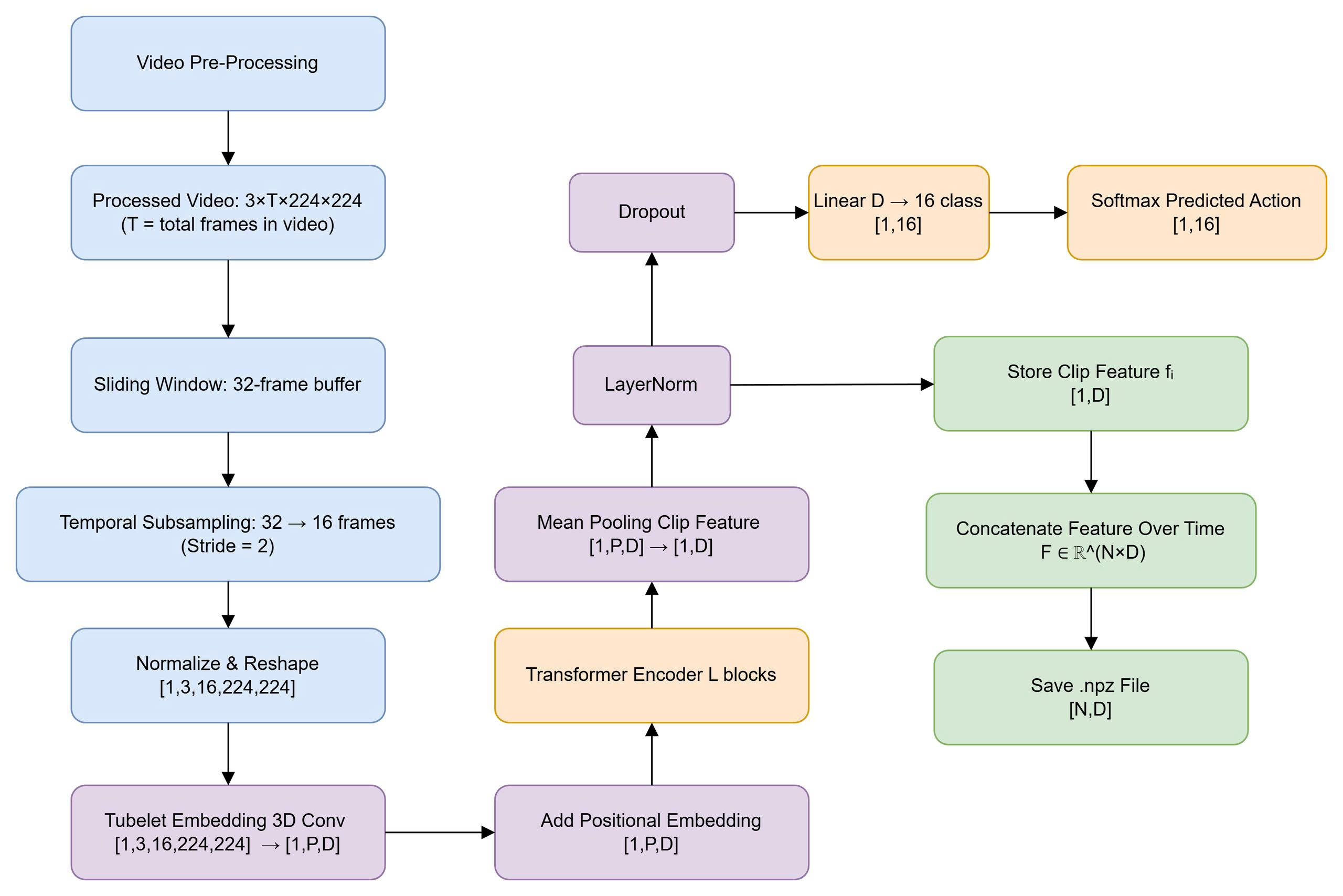}
	\caption{Detailed feature extraction pipeline based on the fine-tuned VideoMAE encoder.}
	\label{fig:Feature_Extraction}
\end{figure}

Feature extraction operates at the chunk level over overlapping 16-frame inputs, mean-pooling token outputs to form a unified video-level feature matrix of size $[N_{\text{chunks}}, D]$ (Figure \ref{fig:Feature_Extraction}, Table \ref{tab:videomae_encoder_config_vertical}).

\begin{table}[h]
	\centering
	\caption{VideoMAE Encoder Configurations for Feature Extraction}
	\label{tab:videomae_encoder_config_vertical}
	\begin{tabular}{lcc}
		\toprule
		\textbf{Configuration} & \textbf{ViT-Base} & \textbf{ViT-Giant} \\
		\midrule
		Patch size & $16 \times 16$ & $14 \times 14$ \\
		Spatial patches & $196\ (14 \times 14)$ & $256\ (16 \times 16)$ \\
		Temporal groups & $8$ & $8$ \\
		Total tubelets ($P$) & $1{,}568$ & $2{,}048$ \\
		Embedding dimension ($D$) & $768$ & $1{,}408$ \\
		Output shape & $[1, 1568, 768]$ & $[1, 2048, 1408]$ \\
		\bottomrule
	\end{tabular}
\end{table}

\subsubsection{Augmented Self-Mask Attention (AMA)}
To capture highly variable driving action durations, the Augmented Self-Mask Attention (AMA) model \cite{zhang2024augmented} enhances standard temporal attention. Given tensors $(Q, K, V) \in \mathbb{R}^{T \times D}$, standard attention is defined as:
\[
S'_a = \mathrm{Softmax}\left(\frac{QK^{\top}}{\sqrt{D}}\right)V. \tag{1}
\]
AMA introduces a permutation-based context mechanism \cite{yang2019xlnet}. A sequence is divided into a preceding segment $\tilde{X}$ and a current segment $X$. Processing $\tilde{X}$ under permutation $\tilde{S}$ yields hidden states $\tilde{H}^{(m-1)}$ that act as a cached memory. The attention update for layer $m$ is:
\[
H^{(m)}_{Z_T} = \mathrm{Softmax}\!\left(Q = H^{(m-1)}_{Z_T}, \; KV = \left[\tilde{H}^{(m-1)},\, H^{(m-1)}_{Z_{\le T}}\right]\right), \tag{2}
\]
where relative positional embeddings preserve the true temporal ordering.

\begin{figure}[h]
	\centering
	\vspace{10pt}
	\includegraphics[scale=0.12]{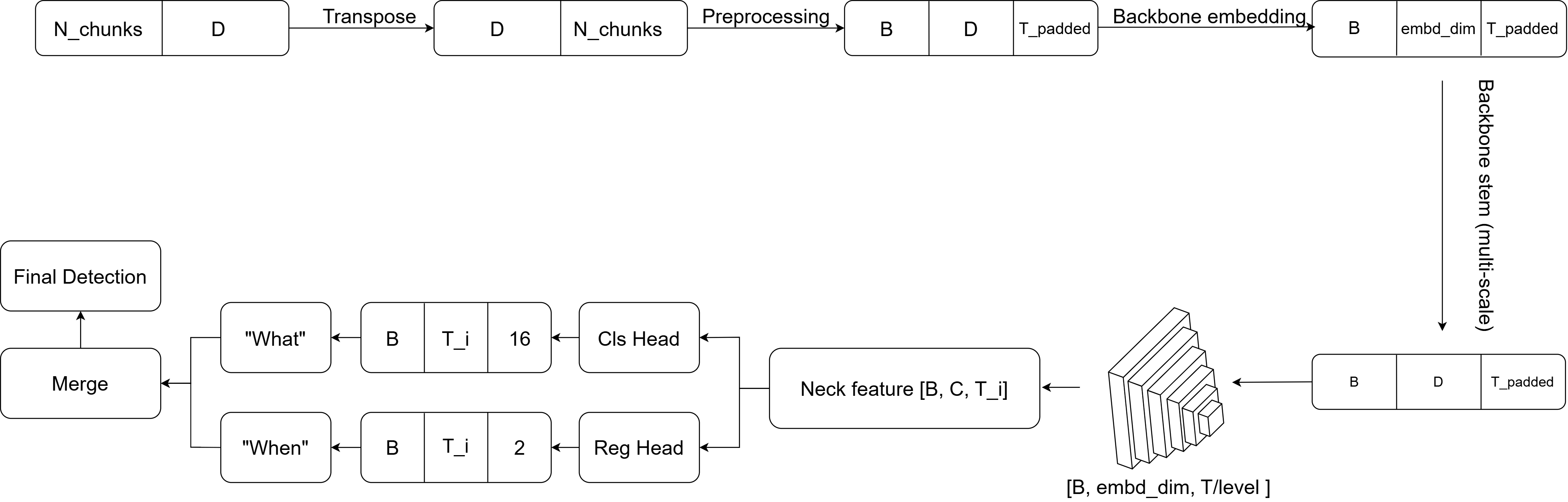}
	\caption{Details of AMA pipeline}
	\label{fig:AMA}
\end{figure}

\begin{figure}[h]
	\centering
	\includegraphics[width=1.0\linewidth]{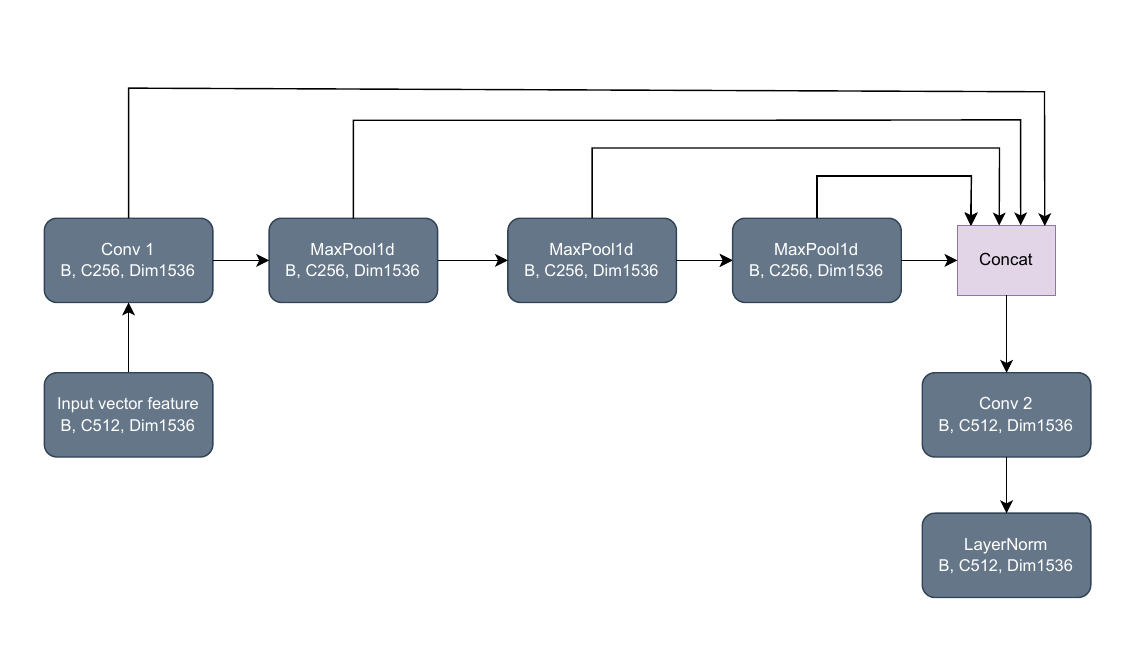}
	\caption{Spatial pyramid pooling fast 1D workflow.}
	\label{fig:SPPF}
\end{figure}

The transposed input matrix $[D, N_{\text{chunks}}]$ passes through a hierarchical backbone, an FPN neck, and parallel heads (Figure \ref{fig:AMA}, Table \ref{tab:ama_config}). To enhance receptive fields beyond the baseline \textit{Identity} neck, we implement a 1D temporal \textbf{Spatial Pyramid Pooling Fast (SPPF)} module \cite{he2015spatial, yolov5}. SPPF compresses channels via a $1 \times 1$ convolution, applies three sequential 1D max-pooling layers ($k=5$, stride=1), and concatenates multi-scale context before a final $1 \times 1$ convolution and LayerNorm (see Figure \ref{fig:SPPF}):
\[
\mathbf{Z} = \mathrm{LN}\bigl(\mathrm{Conv}_2([\mathbf{x}, \mathrm{MP}(\mathbf{x}), \mathrm{MP}^2(\mathbf{x}), \mathrm{MP}^3(\mathbf{x})])\bigr).
\]

Multi-scale features feed a point generator creating temporal anchors. Parallel convolutional heads handle multi-label classification ($\mathbb{R}^{B \times T_i \times N_{\text{classes}}}$) and temporal boundary regression ($\mathbb{R}^{B \times T_i \times 2}$). Predictions are filtered via a pre-NMS threshold ($0.2$) and boundary-decoded:
\[
seg_{\text{left}} = t - \text{offset}_{\text{left}} \times \text{stride}_{\text{fpn}}, \qquad seg_{\text{right}} = t + \text{offset}_{\text{right}} \times \text{stride}_{\text{fpn}}.
\]
Detections are mapped to precise timestamps, filtered by Softmax confidence ($>0.5$), and refined via Non-Maximum Suppression (NMS, threshold=0.5).

\begin{table}[h]
	\centering
	\caption{General Configuration of the AMA Temporal Action Detection Pipeline}
	\label{tab:ama_config}
	\resizebox{\textwidth}{!}{%
		\begin{tabular}{l|l||l|l}
			\hline
			\textbf{Configuration} & \textbf{Value} & \textbf{Configuration} & \textbf{Value} \\ \hline
			Input Feature Shape & $[D, N_{\text{chunks}}]$ & Backbone Type & \texttt{convTransformer} / \texttt{conv} \\
			Dimension ($D$) & $768$ (Base) / $1408$ (Giant) & Neck Configurations & \texttt{identity} / \texttt{sppf} \\
			FPN Levels & $\text{backbone\_arch}[-1] + 1$ & Regression Output & $[B, T_i, 2]$ (offsets) \\
			Classification Output & $[B, T_i, 16]$ & NMS IoU Threshold & $0.5$ \\ \hline
	\end{tabular}}
\end{table}

\subsubsection{Ensemble Model}
To unify redundant or overlapping predictions from the complementary VideoMAE and VideoMAE V2 backbones \cite{zhang2022actionformer, shi2023tridet}, we deploy an ensemble strategy:
\begin{itemize}
	\item \textbf{Step 1 (High-confidence filtering):} Select the top prediction per video ID and action class label to suppress low-quality fragments.
	\item \textbf{Step 2 (Temporal IoU fusion):} Group predictions $S_i^j = \{(start_p, end_p)\}_{p=1}^{N}$ matching in video $i$ and class $j$, merging boundaries unweighted:
	\[
	t_{s,i}^{j} = \frac{1}{N} \sum_{p=1}^{N} start_p , \qquad t_{e,i}^{j} = \frac{1}{N} \sum_{p=1}^{N} end_p . \tag{3}
	\]
	\item \textbf{Step 3 (Score-weighted fusion):} Refine boundaries by prioritizing predictions with higher confidence weights:
	\[
	t_{s,i}^{j} = \frac{\sum_{p=1}^{N} start_p \cdot score_p}{\sum_{p=1}^{N} score_p}, \qquad t_{e,i}^{j} = \frac{\sum_{p=1}^{N} end_p \cdot score_p}{\sum_{p=1}^{N} score_p}. \tag{4}
	\]
\end{itemize}

\section{Results and Discussion}
\label{sec:resultsdiscussion}
\subsection{Results and Discussion}
\subsubsection{Training Setup and Hyperparameters}
The training pipeline is executed in two independent stages on a server equipped with an NVIDIA RTX A5000 GPU (24~GB VRAM) and 64~GB RAM. 

\paragraph{Feature Extraction (VideoMAE Fine-tuning):} Pretrained models are fine-tuned for 45 epochs on the AICity2024 Track~3 dataset using AdamW with an initial learning rate of $1\times10^{-3}$, betas $(0.9, 0.999)$, weight decay $0.05$, and gradient clipping at $5.0$. A cosine annealing scheduler is applied with a 5-epoch linear warmup. Layer-wise decay is set to $0.75$ (ViT-B) and $0.80$ (ViT-G). Inputs are resized to $224 \times 224$ (16 frames per clip, stride 4) with a batch size of 1. Regularization includes drop path rates of $0.1$ (ViT-B) / $0.2$ (ViT-G) and classification head dropouts of $0.0$ (ViT-B) / $0.5$ (ViT-G).

\paragraph{Action Localization (AMA Module):} The AMA module is optimized for 30 epochs over frozen backbone features using AdamW with a learning rate of $1\times10^{-4}$ and weight decay $0.05$. The cosine scheduler includes 5 warmup epochs. Sequences are capped at a maximum length of 1536 frames with a feature stride of 16. Optimization targets a combined loss consisting of Sigmoid Focal loss for classification, 1D DIoU loss for temporal regression, and an auxiliary action localization loss (weight $0.2$). Post-processing retains the top 5000 predictions above a 0.2 confidence threshold before multi-class NMS.

\subsubsection{Backbone and AMA Training Dynamics}
\begin{figure}[h]
	\centering
	\includegraphics[width=0.32\linewidth]{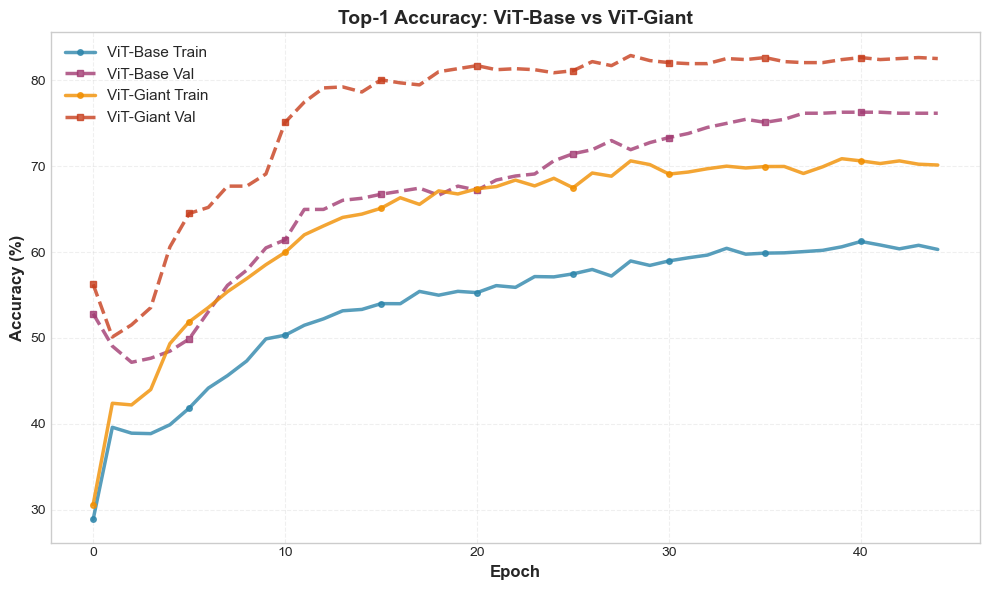}\hfill
	\includegraphics[width=0.32\linewidth]{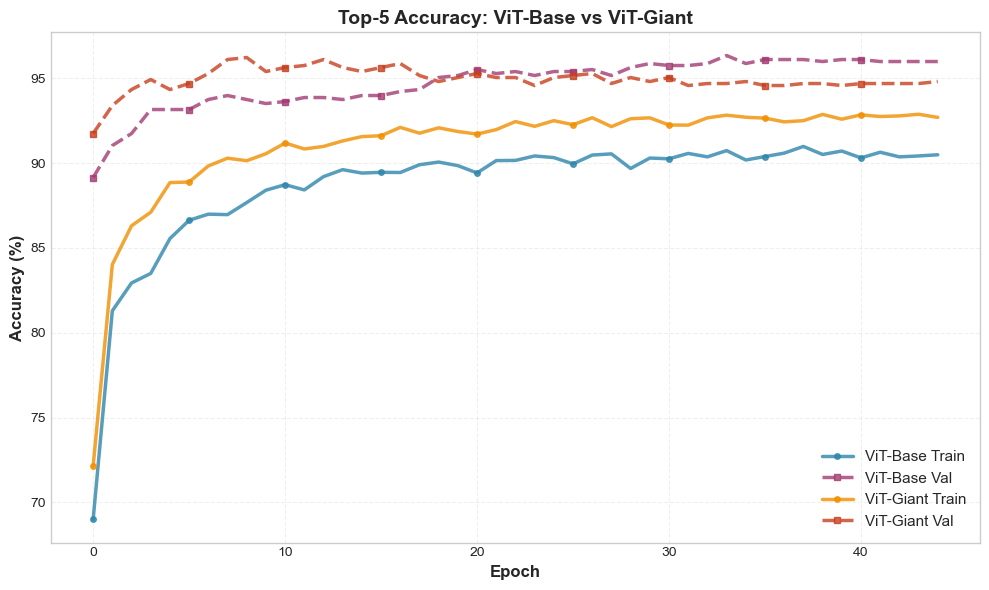}\hfill
	\includegraphics[width=0.32\linewidth]{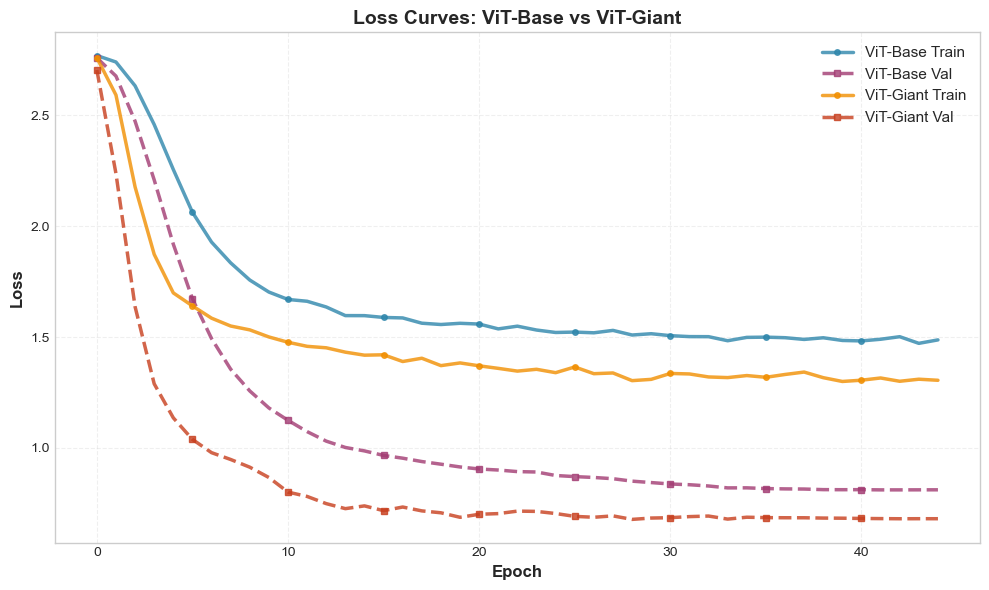}
	\caption{Training and validation Top-1 accuracy (left), Top-5 accuracy (center), and loss (right) curves.}
	\label{fig:vit_curves}
\end{figure}

The VideoMAE backbones exhibit distinct optimization behavior according to model scale. ViT-Base achieves stable convergence smoothly over 45 epochs, while ViT-Giant exhibits higher validation metrics early on due to its massive parameter capacity as shown in Figures \ref{fig:vit_curves}. 

\begin{figure}[h]
	\centering
	\begin{subfigure}[t]{0.48\textwidth}
		\centering\includegraphics[width=\textwidth]{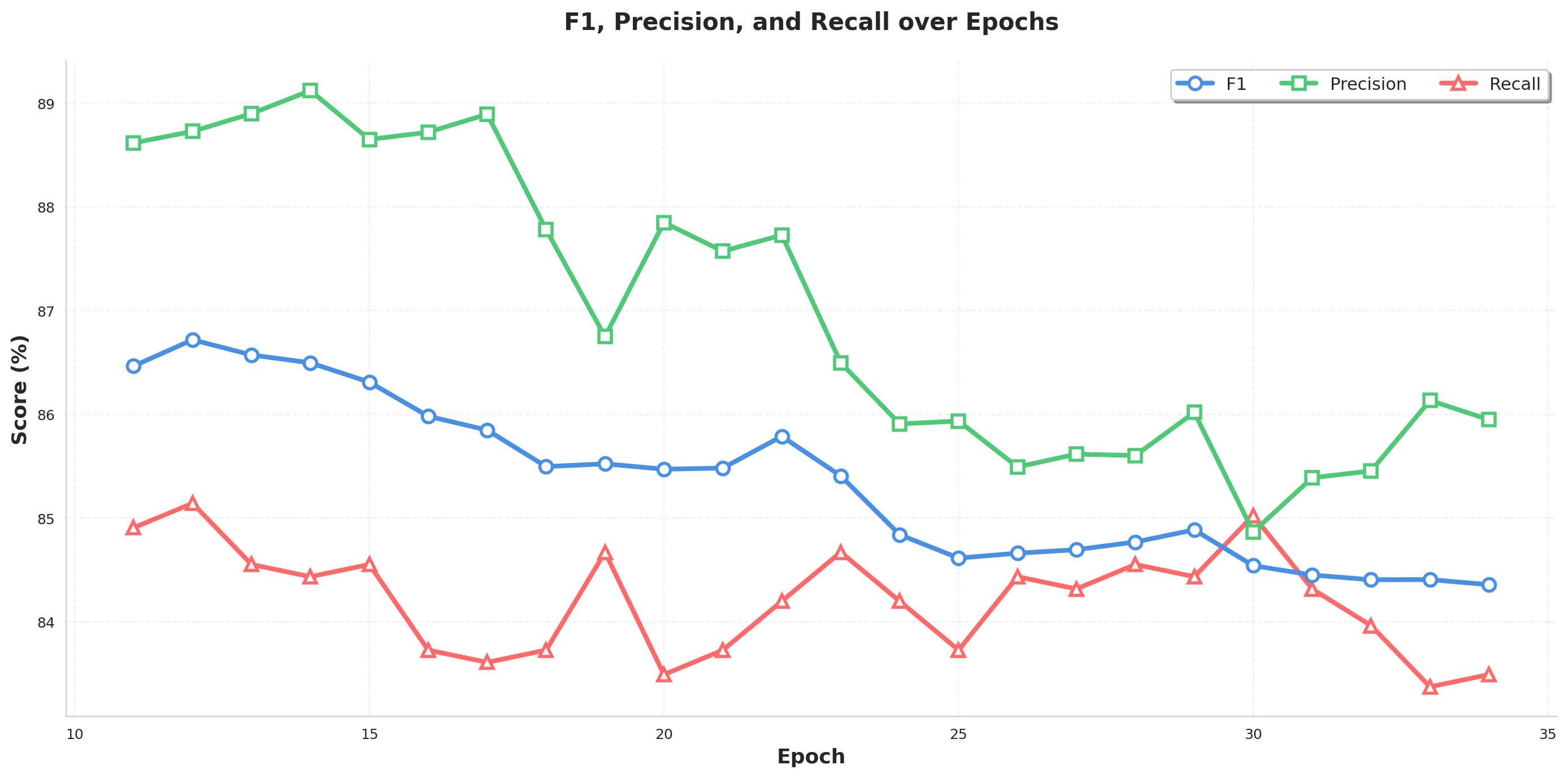}
		\caption{ViT-Base + Identity}
	\end{subfigure}\hfill
	\begin{subfigure}[t]{0.48\textwidth}
		\centering\includegraphics[width=\textwidth]{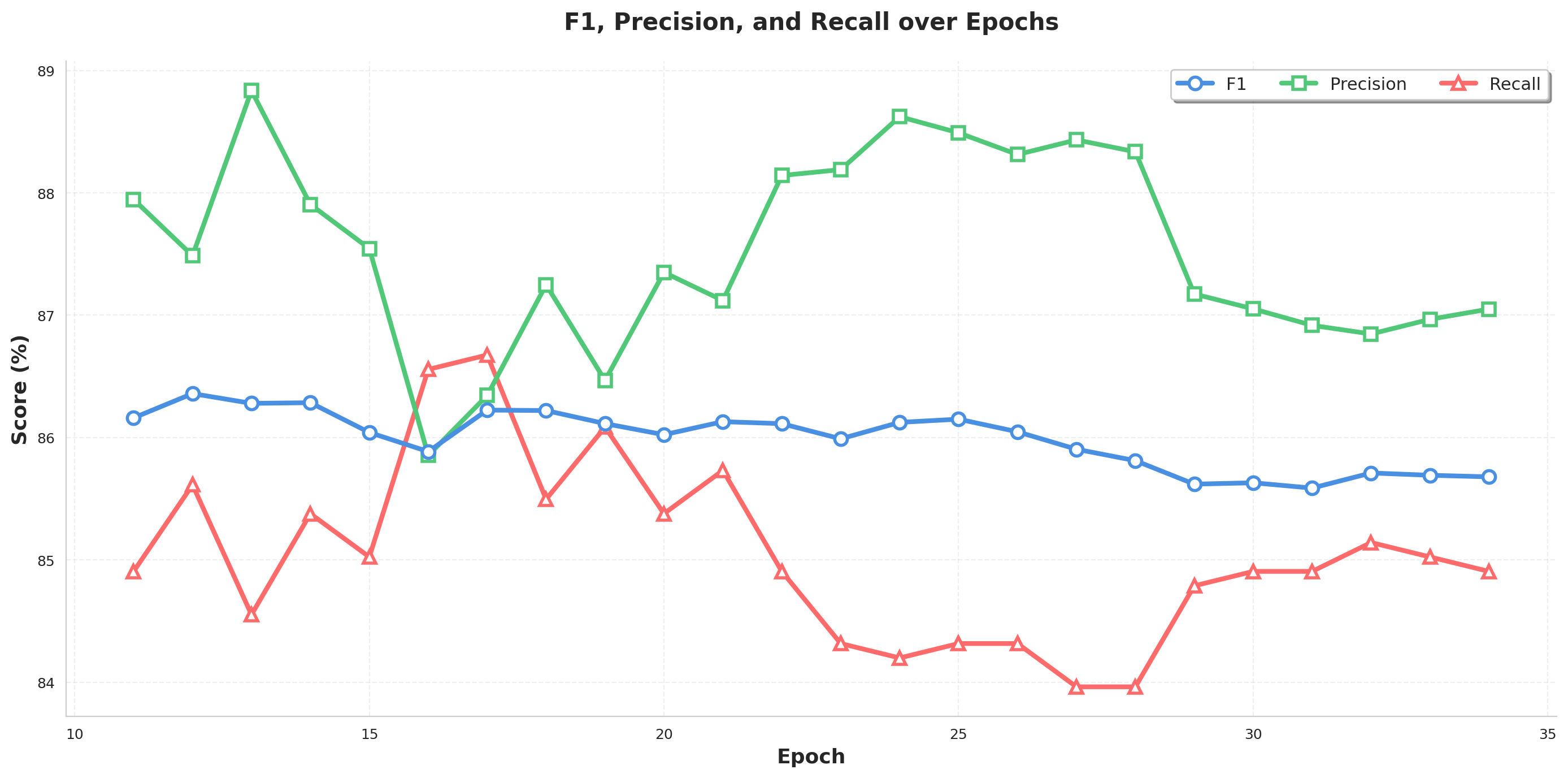}
		\caption{ViT-Base + SPPF}
	\end{subfigure}
	\caption{AMA training dynamics on ViT-Base features using Identity and SPPF necks.}
	\label{fig:ama_vitb_combined}
\end{figure}

\begin{figure}[h]
	\centering
	\begin{subfigure}[t]{0.48\textwidth}
		\centering\includegraphics[width=\textwidth]{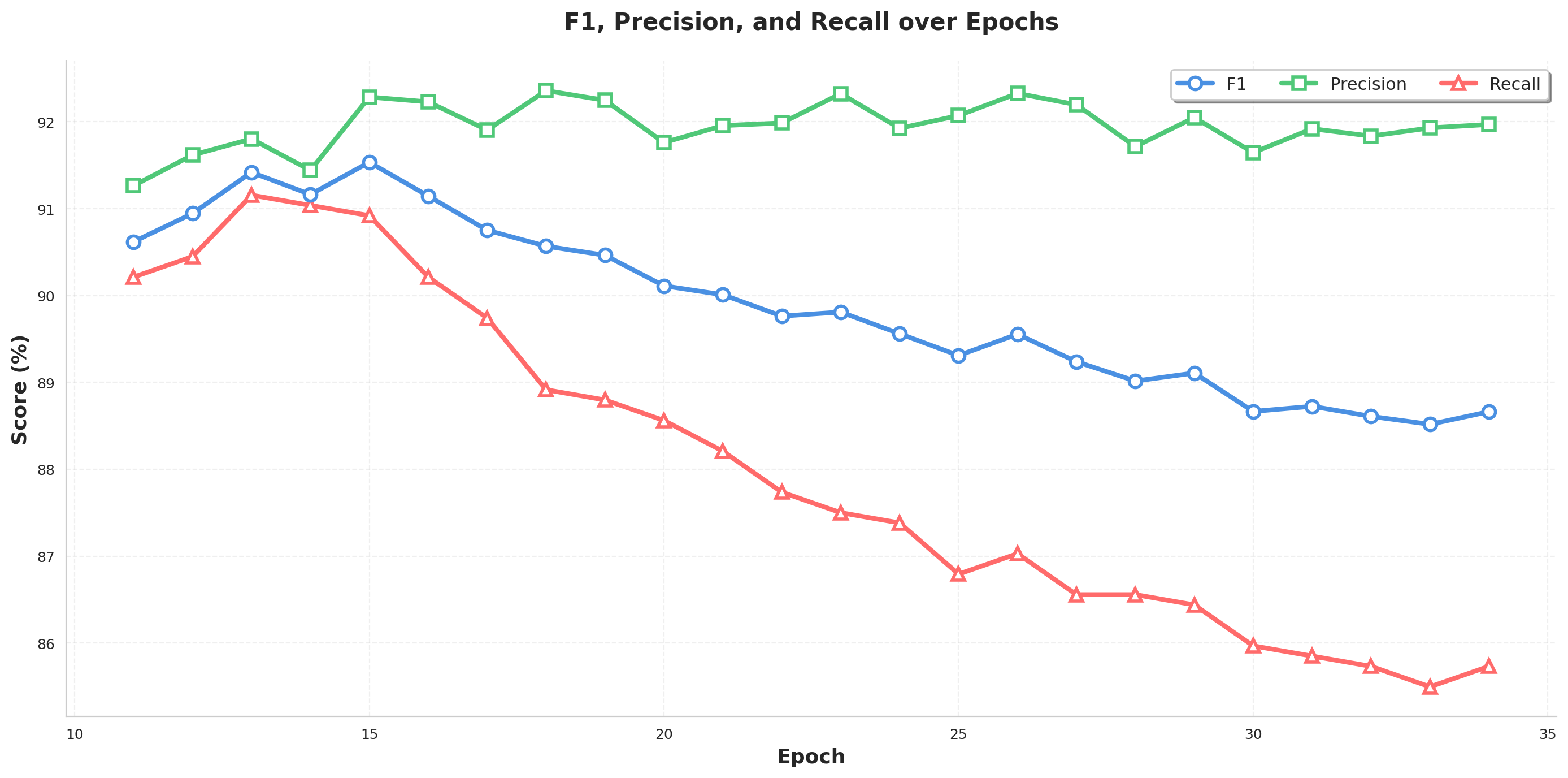}
		\caption{ViT-Giant + Identity}
	\end{subfigure}\hfill
	\begin{subfigure}[t]{0.48\textwidth}
		\centering\includegraphics[width=\textwidth]{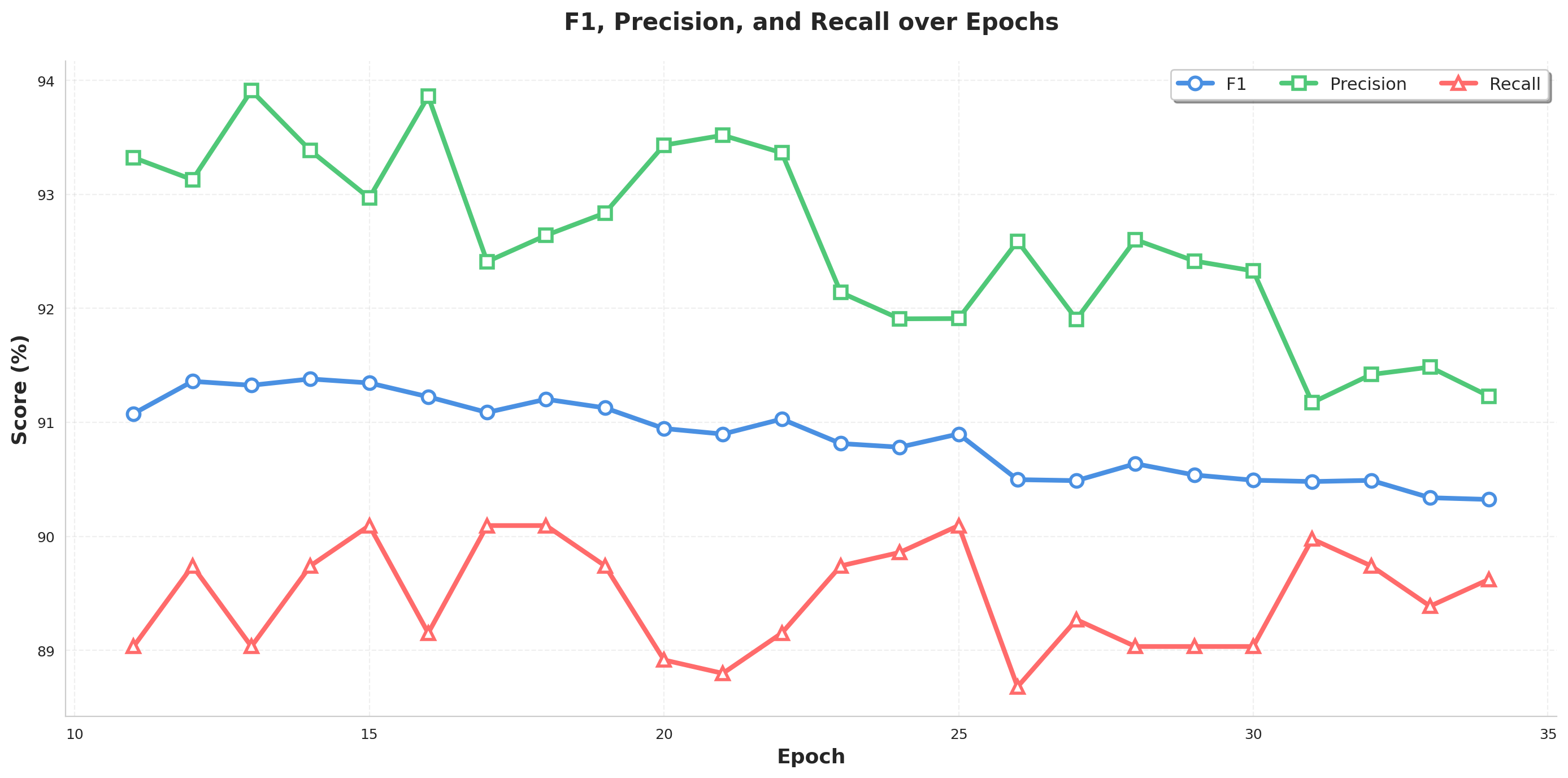}
		\caption{ViT-Giant + SPPF}
	\end{subfigure}
	\caption{AMA training dynamics on ViT-Giant features using Identity and SPPF necks.}
	\label{fig:ama_vitg_combined}
\end{figure}

For the localization stage, the AMA module reaches rapid convergence due to the pre-computed representations. Figures~\ref{fig:ama_vitb_combined} and \ref{fig:ama_vitg_combined} contrast the behavior of the base Identity neck against our proposed SPPF neck. The SPPF configuration drives higher peak precision and uniform class-wise mAP by expanding the multi-scale temporal context, which yields marked performance gains on long-duration or complex actions (e.g., reaching behind, picking items from floor).

\begin{figure}[h]
	\centering
	\begin{subfigure}[t]{0.48\textwidth}
		\centering\includegraphics[width=\textwidth]{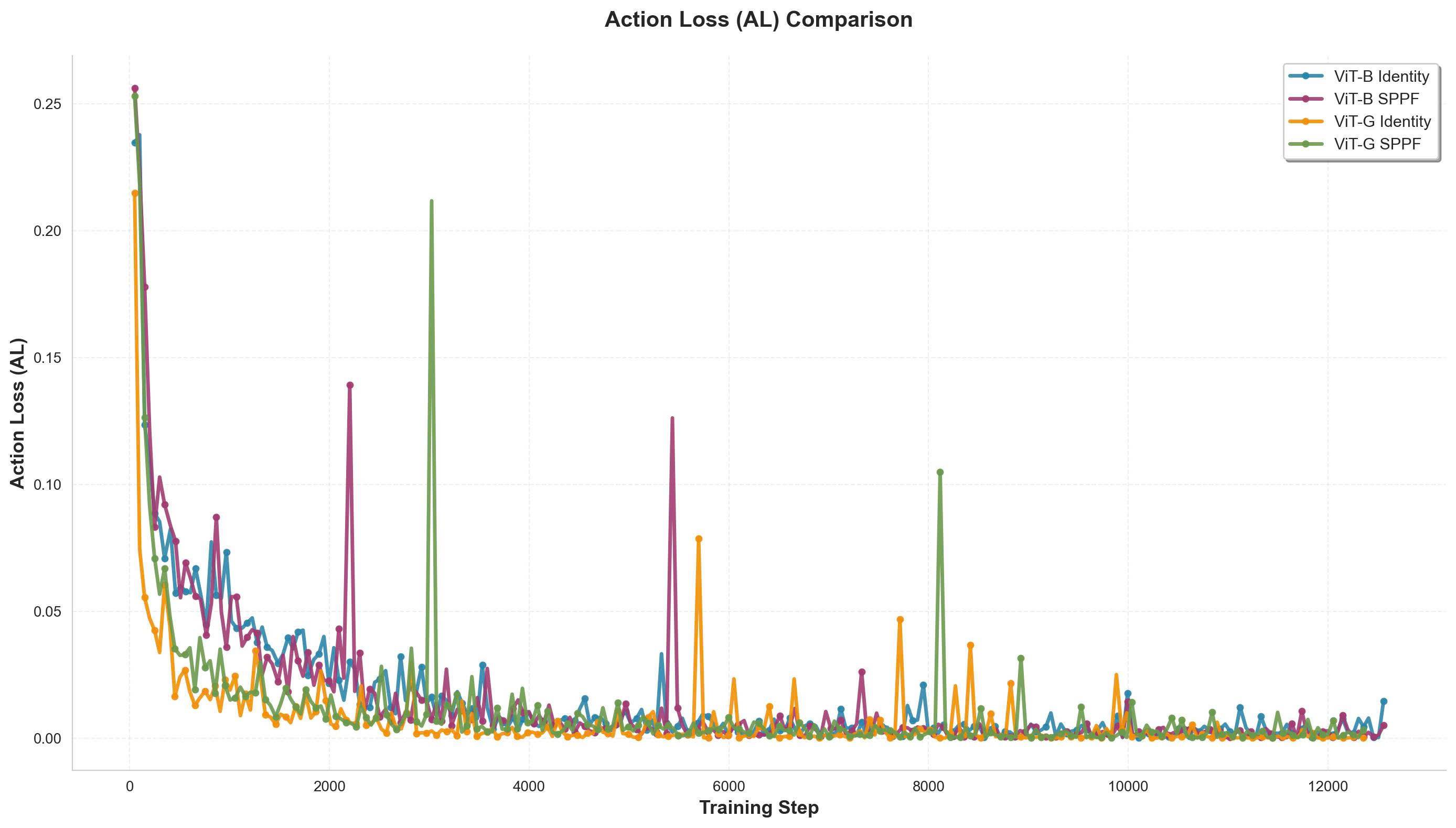}
		\caption{Action Loss}
	\end{subfigure}\hfill
	\begin{subfigure}[t]{0.48\textwidth}
		\centering\includegraphics[width=\textwidth]{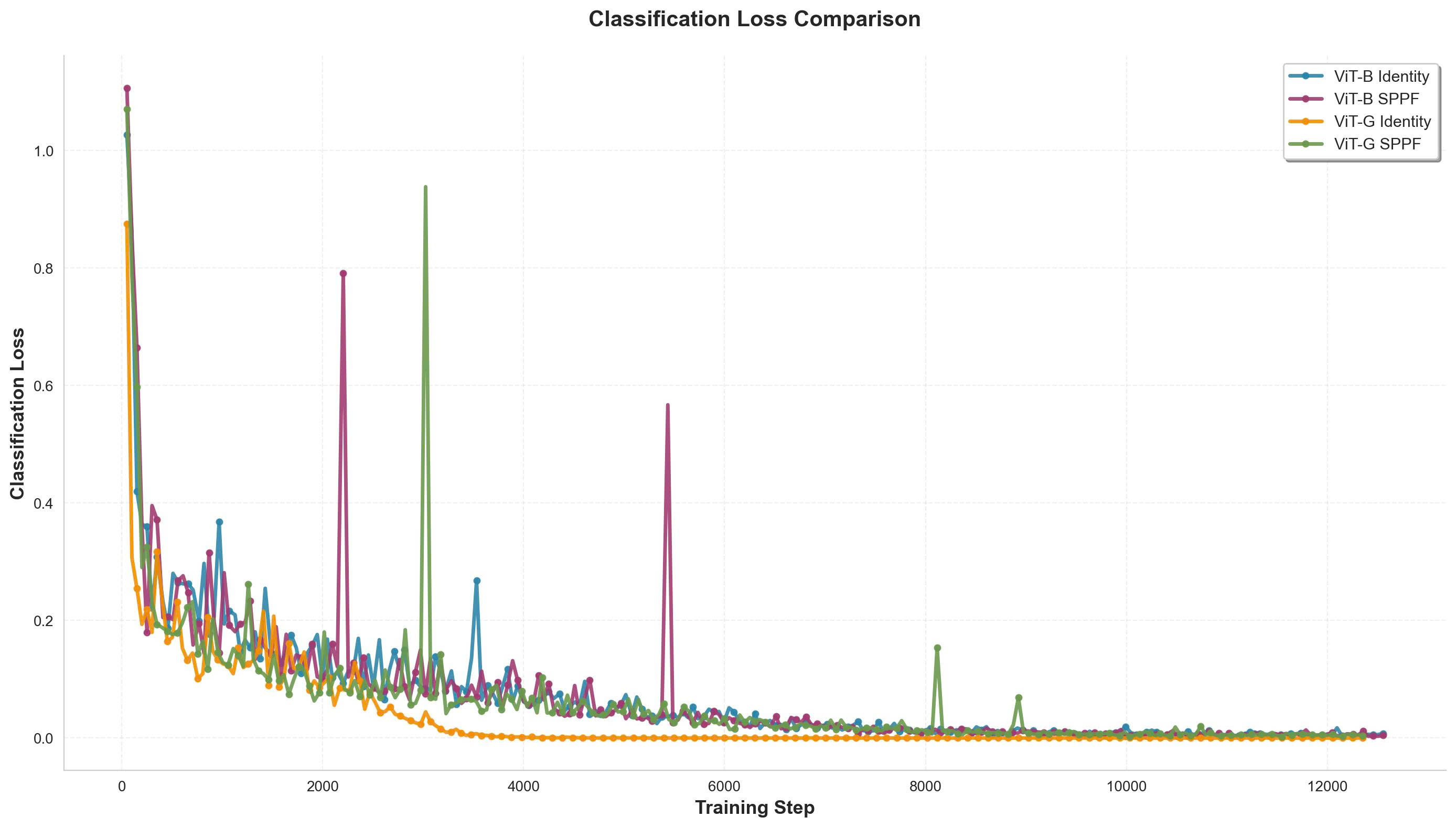}
		\caption{Classification Loss}
	\end{subfigure}
	\caption{Step-wise loss trajectory comparison between model configurations.}
	\label{fig:loss_comparison}
\end{figure}

As shown in Figure~\ref{fig:loss_comparison}, Identity models maintain a monotonic, smooth logarithmic decay. Conversely, the SPPF models induce localized loss spikes (e.g., around steps 2,200 and 5,500), reflecting active parameter space exploration that escapes local minima before rapidly reconverging to a superior overall optimum.

\subsubsection{Main Quantitative Results}
Table~\ref{tab:feature_extraction_results} details the core accuracy-efficiency trade-offs of the backbones. ViT-Giant yields substantial improvements, raising the final test Top-1 accuracy to $88.09\%$ ($+5.54\%$ over ViT-Base). However, this accuracy boost scales the parameter footprint up by over \textbf{11$\times$} and requires $1584.06$ GFLOPs per segment, increasing overall training time from roughly 25 hours to over 131 hours.

\begin{table}[h]
	\centering
	\caption{Comparison of Feature Extraction Results for ViT-Base and ViT-Giant}
	\label{tab:feature_extraction_results}
	\small
	\begin{tabular}{lcc}
		\toprule
		\textbf{Metric} & \textbf{ViT-Base} & \textbf{ViT-Giant} \\
		\midrule
		Best Train Acc@1 / Acc@5 & 61.24\% / 90.98\% & 70.88\% / 92.88\% \\
		Best Val Acc@1 / Acc@5 & 76.30\% / 96.34\% & 82.90\% / 96.23\% \\
		Final Test Acc@1 / Acc@5 & 82.55\% / 98.47\% & 88.09\% / 99.06\% \\
		\midrule
		Model FLOPs / Parameters & 101.85 GFLOPs / 86.2M & 1584.06 GFLOPs / 1011.6M \\
		Total Training / Avg Epoch Time & 24.95 hours / 31.35 mins & 131.83 hours / 169.21 mins \\ 
		\bottomrule
	\end{tabular}
\end{table}

Tables~\ref{tab:ama_vit_results_base} and \ref{tab:best_epoch_results} benchmark localization results for different neck options across runs. The integration of SPPF consistently outperforms the baseline Identity configurations. On the ViT-Base features, SPPF raises the average mAP by nearly $2\%$. This effect is amplified when scaling to ViT-Giant features, where SPPF delivers a mean mAP of $92.01\%$ while significantly suppressing cross-run variance ($\sigma$ dropping from $1.68$ down to $0.46$).

\begin{table}[h]
	\centering
	\caption{Mean $\pm$ Std AMA localization performance cross-comparison.}
	\label{tab:ama_vit_results_base}
	\small
	\begin{tabular}{lcccc}
		\toprule
		\textbf{Configuration} & \textbf{avg\_mAP} & \textbf{F1} & \textbf{Precision} & \textbf{Recall} \\
		\midrule
		ViT-Base + Identity & 86.26\% $\pm$ 0.77 & 85.34\% $\pm$ 0.79 & 87.01\% $\pm$ 1.42 & 84.22\% $\pm$ 0.51 \\
		ViT-Base + SPPF & \textbf{88.05\% $\pm$ 0.67} & \textbf{85.99\% $\pm$ 0.24} & \textbf{87.53\% $\pm$ 0.79} & \textbf{85.04\% $\pm$ 0.73} \\
		\midrule
		ViT-Giant + Identity & 89.13\% $\pm$ 1.68 & 89.89\% $\pm$ 0.97 & 91.95\% $\pm$ 0.29 & 88.04\% $\pm$ 1.90 \\
		ViT-Giant + SPPF & \textbf{92.01\% $\pm$ 0.46} & \textbf{90.87\% $\pm$ 0.35} & \textbf{92.58\% $\pm$ 0.82} & \textbf{89.46\% $\pm$ 0.46} \\
		\bottomrule
	\end{tabular}
\end{table}

\begin{table}[h]
	\centering
	\caption{Best--epoch AMA performance comparison.}
	\label{tab:best_epoch_results}
	\small
	\begin{tabular}{lcccc}
		\toprule
		\textbf{Model} & \textbf{mAP} & \textbf{F1} & \textbf{Precision} & \textbf{Recall} \\
		\midrule
		ViT-Base + Identity & 87.94\% & 86.46\% & 88.62\% & 84.91\% \\
		ViT-Base + SPPF     & 89.10\% & 86.16\% & 87.95\% & 84.91\% \\
		ViT-Giant + Identity & 92.12\% & 90.62\% & 91.27\% & \textbf{90.21}\% \\
		ViT-Giant + SPPF     & \textbf{92.67}\% & \textbf{91.33}\% & \textbf{93.91}\% & 89.03\% \\
		\bottomrule
	\end{tabular}
\end{table}

\subsection{Discussion and Interpretation}
The experimental results demonstrate a stark trade-off between model capacity and computational efficiency. The billion-parameter ViT-Giant backbone captures nuanced spatiotemporal representations far superior to ViT-Base, proving that complex driver actions involving subtle motions and occlusions benefit from massive model scales. Furthermore, the structural enhancement of adding a Spatial Pyramid Pooling Fast (SPPF) neck consistently outperforms the baseline Identity design. Because in-cabin behaviors vary from prolonged (e.g., phone usage) to fleeting (e.g., glancing), the SPPF multi-scale context aggregation allows the network to adapt to varying temporal durations. 

However, class-wise analysis exposes a fundamental representation bottleneck: while gross motor actions involving distinct body shifts achieve high precision, the model struggles with fine-grained, localized behaviors such as \textit{Yawning}, \textit{Talking}, or \textit{Singing}. These activities rely heavily on facial micro-expressions and mouth movements, which are easily overshadowed by global body features in a downsampled, video-only transformer architecture.

\paragraph{Practical Implications:}
These findings suggest a bifurcated deployment strategy. The compact ViT-Base framework remains the logical baseline for resource-constrained, real-time edge devices, whereas the heavy ViT-Giant framework is ideal for server-side batch diagnostics or as a teacher network in a knowledge distillation pipeline. Additionally, the clear benefits of the SPPF neck imply that future in-cabin action localization models should prioritize hierarchical temporal context pooling. Finally, the visual ambiguity among facial activities indicates that vision-only architectures reach an operational ceiling, validating the necessity of multi-modal features for high-precision safety auditing.

\subsection{Limitations and Future Works}
\paragraph{Limitations:}
\begin{itemize}
	\item \textbf{Domain Vulnerability:} Optimization and evaluation were restricted to a single benchmark dataset, introducing potential sensitivity to domain shift under unseen cabin geometries or illumination profiles.
	\item \textbf{Hardware Constraints:} Memory limits forced a training batch size of one during backbone fine-tuning, which potentially introduced gradient variance during backpropagation.
	\item \textbf{Deployment Footprint:} The heavy computational demand of ViT-Giant prevents real-time, on-vehicle deployment on commercial automotive embedded systems.
	\item \textbf{Fine-Grained Ambiguity:} The vision-only paradigm fails to robustly decouple visually identical but semantically distinct categories lacking distinct body articulation (e.g., singing vs. talking).
\end{itemize}

\paragraph{Future Works:}
To improve system viability, future research will explore Unsupervised Domain Adaptation (UDA) across heterogeneous datasets to combat domain shift. We plan to apply knowledge distillation techniques to compress the representations of our ViT-Giant model into a lightweight, edge-deployable student network. Lastly, we aim to extend the framework into a multimodal pipeline by incorporating synchronized in-cabin audio processing, leveraging acoustic features to accurately resolve ambiguities between yawning, singing, and talking.

\section{Conclusion}
\label{sec:conclusion}
\subsection{Summary of Contributions}
This study presented a robust two-stage framework for temporal action localization in driver monitoring systems, combining the spatiotemporal representational power of Masked Autoencoders (VideoMAE) with an Augmented Self-Mask Attention (AMA) module. By decoupling representation learning from temporal boundary regression, we systematically benchmarked the impact of model capacity using ViT-Base and ViT-Giant backbones. The core contribution lies in this extensive self-supervised evaluation alongside the validation of a multi-scale pooling mechanism within the localization head, bypassing the limitations of traditional supervised pretraining to capture richer driving-context dynamics.

\subsection{Key Findings}
\begin{itemize}
	\item \textbf{Capacity Dominance:} The billion-parameter ViT-Giant backbone consistently outperformed the ViT-Base variant, proving that high-capacity models are crucial for capturing subtle, occluded, and fine-grained spatiotemporal driving behavior cues.
	\item \textbf{Contextual Receptive Fields:} The integration of a 1D Spatial Pyramid Pooling Fast (SPPF) neck significantly improved localization boundary precision by effectively aggregating multi-scale context to accommodate heterogeneous action durations.
	\item \textbf{Efficiency Bottlenecks:} A stark accuracy-efficiency trade-off remains; while ViT-Giant maximizes detection robustness, its high GFLOP footprint limits real-time deployment on edge hardware compared to the compact ViT-Base framework.
	\item \textbf{Visual Operational Ceiling:} Fine-grained facial and physiological actions (e.g., yawning vs. talking) remain visually ambiguous, underscoring a baseline limitation in vision-only streams.
\end{itemize}

\subsection{Closing Remarks}
In conclusion, this work bridges the gap between academic vision research and practical intelligent transportation deployment. Rather than evaluating isolated metrics, we systematically mapped accuracy against computational overhead to guide embedded system integration. 

The resulting architecture provides a highly extensible baseline for in-cabin behavioral auditing and commercial traffic safety compliance. Coupled with our web-based demonstration system, this framework operationalizes deep transformer-based temporal action localization, laying a reliable foundation for next-generation intelligent driver assistance systems (ADAS).

\section*{Declaration of conflicting interest}
The authors declared no potential conflicts of interest with respect to the research, authorship, and/or publication of this article.


\section*{Data availability statement}
Data supporting the findings of this article are publicly available at \url{https://www.aicitychallenge.org/2024-data-and-evaluation/}.

\section*{ORCID iD}
Thi-Thu-Hien Pham: \url{https://orcid.org/0000-0001-5808-3214}\\
Thanh-Hai Le: \url{https://orcid.org/0000-0002-3212-3940}

\bibliographystyle{ieeetr}  
\bibliography{references}  

\end{document}